%% file: neurips_2026_final.tex
\newcommand\FINAL[1]{}

\providecommand\FINAL[1]{#1}
\newcommand\PASFINAL[1]{#1}
\FINAL{\renewcommand\PASFINAL[1]{}}

\newcommand\todoinlineD[1]{\FINAL{}}

\newcommand\ICALPSHORTER[1]{}
\newcommand\ICALPDETAILED[1]{}
\newcommand\ICALPSHORTERJOHANNE[1]{}

 \newcommand\CONFIGURATIONSMT{\mathcal{C}_{M}}

 \newcommand{\step}{\operatorname{step}}
 \newcommand{\enc}{\operatorname{enc}}
  \newcommand{\Cert}{\varepsilon}     %

   \newcommand{\sLR}{\sigma_\alpha}            %
   \newcommand{\AccODE}{R_{\mathrm{acc}}^{\mathrm{ODE}}}
   \newcommand{\InitODE}{R_{\mathrm{init}}^{\mathrm{ODE}}}

  \newcommand{\HALT}{\operatorname{HALT}}
 
 \newcommand\dist{dist}

\newcommand\Init{\operatorname{Init}}
\newcommand\Acc{\operatorname{Acc}}

\newcommand\NP{\operatorname{NP}}
 \newcommand{\coNP}{\operatorname{coNP}}
   \newcommand{\PTIME}{\operatorname{PTIME}}

\newcommand\NEXP{\operatorname{NEXP}}
\newcommand\EXPTIME{\operatorname{EXPTIME}}

\newcommand\SAT{\operatorname{SAT}}
\newcommand\CNF{\operatorname{CNF}}
\newcommand{\promiseNP}{\operatorname{promiseNP}}

\newcommand\RE{\operatorname{RE}}

\newcommand\RP{\mathbb{R}_{\ge 0}}

\documentclass{article}

\usepackage[main, final]{neurips_2026}

 \usepackage{makecell}  %

\usepackage[utf8]{inputenc} %
\usepackage[T1]{fontenc}    %
\usepackage{hyperref}       %
\usepackage{url}            %
\usepackage{booktabs}       %
\usepackage{amsfonts}       %
\usepackage{nicefrac}       %
\usepackage{microtype}      %
\usepackage{xcolor}         %
\usepackage{adjustbox}

\author{%
  David S.~Hippocampus\thanks{Use footnote for providing further information
    about author (webpage, alternative address)---\emph{not} for acknowledging
    funding agencies.} \\
  Department of Computer Science\\
  Cranberry-Lemon University\\
  Pittsburgh, PA 15213 \\
  \texttt{hippo@cs.cranberry-lemon.edu} \\
}

\usepackage{amsmath,amssymb,amsthm}

\newtheorem{theorem}{Theorem}
\newtheorem{lemma}[theorem]{Lemma}
\newtheorem{proposition}[theorem]{Proposition}
\newtheorem{corollary}[theorem]{Corollary}

\theoremstyle{definition}
\newtheorem{definition}[theorem]{Definition}

\newtheorem*{theorem*}{Theorem}
\newtheorem*{corollary*}{Corollary}

\theoremstyle{remark}
\newtheorem{remark}[theorem]{Remark}

\usepackage{cleveref}
\AddToHook{env/lemma/begin}{\crefalias{theorem}{lemma}}
\AddToHook{env/proposition/begin}{\crefalias{theorem}{proposition}}
\AddToHook{env/corollary/begin}{\crefalias{theorem}{corollary}}
\AddToHook{env/definition/begin}{\crefalias{theorem}{definition}}
\AddToHook{env/remark/begin}{\crefalias{theorem}{remark}}

\usepackage{versions}

\excludeversion{ICALP}
\excludeversion{Skipconnections}
\excludeversion{EXTRAPOLE} %

\newcommand\Netnotation[4]{\mathcal{N} \mathcal{N}^{#1}_{#2,#3,#4}}

\newcommand\Netnotatione[3]{\mathcal{N} \mathcal{N}^{#1}_{#2,#3,*}}

\newcommand\Err{\textrm{err}}
\newcommand\NNErr{\NN_{\Err}}
\newcommand\lead{\textrm{lead}}
\newcommand\shift{\textrm{shift}}

\newcommand\NNlead{\NN_{\lead}}
\newcommand\NNshift{\NN_{\shift}}

\newcommand\NNT[1]{\NN_{T(#1)}}

\newcommand{\myrect}[5]{%
    \draw (#1,#2) rectangle (#1+#3,#2+#4);
    \node at (#1 + #3/2, #2 + #4/2) {\small #5};
}

\newcommand{\myrectrotate}[5]{%
    \draw (#1,#2) rectangle (#1+#3,#2+#4);
    \node[rotate=90] at (#1 + #3/2, #2 + #4/2) {\small #5};
}

\usepackage{url}
\PassOptionsToPackage{dvipsnames}{xcolor}

\theoremstyle{plain}

\crefname{lemma}{Lemma}{Lemmas}
\Crefname{lemma}{Lemma}{Lemmas}
\crefname{theorem}{Theorem}{Theorems}
\crefname{proposition}{Proposition}{Propositions}
\crefname{corollary}{Corollary}{Corollaries}
\crefname{definition}{Definition}{Definitions}
\crefname{remark}{Remark}{Remarks}

\usepackage{amscd}
\usepackage{url}
\usepackage{multicol}

\newcommand\finprovisoire{
	\bibliographystyle{plain}
	\bibliography{perso,bournez,literatur,Johanne}
	\end{document}
}

\usepackage{amsmath}
\usepackage{amsthm}
\usepackage{graphicx}
\usepackage{algpseudocode}
\FINAL{\usepackage[colorinlistoftodos]{todonotes}}
\PASFINAL{\newcommand\todo[1]{}}

\newcommand\R{\mathbb{R}}

\newcommand\N{\mathbb{N}}

\newcommand\onethreefouradic{\mathbb{\mathbb{DD}}_{1,3}}

\newcommand\lin{aff}

\usepackage{tikz,xcolor,hyperref}
\usepackage{pgfplots}
\pgfplotsset{compat=1.18}

\usetikzlibrary{arrows.meta}

\definecolor{lime}{HTML}{A6CE39}
\DeclareRobustCommand{\orcidicon}{
	\begin{tikzpicture}
	\draw[lime, fill=lime] (0,0) 
	circle [radius=0.16] 
	node[white] {{\fontfamily{qag}\selectfont \tiny ID}};
	\draw[white, fill=white] (-0.0625,0.095) 
	circle [radius=0.007];
	\end{tikzpicture}
	\hspace{-2mm}
}

\foreach \x in {A, ..., Z}{%
	\expandafter\xdef\csname orcid\x\endcsname{\noexpand\href{https://orcid.org/\csname orcidauthor\x\endcsname}{\noexpand\orcidicon}}
}

\newcommand\ourshortitle{
Iterated Narrow Networks: Turing Completeness\\
       at Width 4 on Convex Input Spaces\\
       and the Complexity of Verification}
\newcommand\ourtitle{\ourshortitle
}
\title{\ourtitle}

\author{%
  Olivier Bournez \\
  Institut Polytechnique de Paris, LIX \\
  91128 Palaiseau, France \\
  \texttt{olivier.bournez@lix.polytechnique.fr} \\
  \And
  Johanne Cohen \\
  LISN, CNRS, Universit\'e Paris-Saclay \\
  91190 Gif-sur-Yvette, France \\
  \texttt{johanne.cohen@lisn.fr} \\
  \And
  Adrian Wurm \\
  Computer Science Institute \\
  BTU Cottbus-Senftenberg \\
  D-03046 Cottbus, Germany \\
  \texttt{wurm@b-tu.de}
}

\newcommand\NN{{N}}

\newcommand\ReLU{\operatorname{ReLU}}

\begin{document}

\maketitle

\begin{abstract}
We study the iterates of a fixed $\ReLU$ network $N$ of width $4$, that is,
the discrete-time system $x_{k+1} = N(x_k)$ on $[0,1]^4$. Equivalently,
$x_k$ is the output of a weight-tied network made of $k$ copies of $N$, or of a
recurrent network with no external input, started at $x_0$. We prove that
every deterministic Turing machine $M$ admits such a network $N_M$. The
network can be constructed in polynomial time from $M$. Each iteration has
depth $\mathrm{poly}(|M|)$. The iterates of $N_M$ simulate $M$ exactly on
every encoded tape. The number of iterations is proportional to the
running time of $M$. Moreover, the fourth coordinate irreversibly flags
every point of the convex cube $[0,1]^4$ whose simulation reads an invalid digit. This flag is the certificate that makes hardness reductions sound on
\emph{convex} input regions, the regime of practical verifiers. Convex
regions were previously inaccessible because of a continuity obstruction
inherent to $\ReLU$ networks. The correspondence has
three consequences. First, reachability for iterated width-$4$
$\ReLU$ networks spans the full complexity hierarchy along the
single axis of the time bound: it is $\NP$-complete for a unary
bound, $\NEXP$-complete for a binary bound and $\RE$-complete
(undecidable) with no bound. Second, two extra neurons of
width suffice to extend the hardness to every common smooth activation: at
width $6$, the robust verification problem (a promise problem with
$\delta/2\delta$ gap) is $\NP$-complete for $\tanh$, sigmoid, GELU, Swish,
ELU, and any polynomial-time-computable continuous non-polynomial
activation with a polynomial-time-computable Lipschitz constant. Third,
the whole spectrum transfers to discretized Neural ODEs in dimension~$5$.
\end{abstract}

\section{Introduction}
\label{sec:intro}
 
We study the iterates of a fixed $\ReLU$ network of constant width:
the discrete-time system
\[
   x_{k+1} \;=\; N(x_k), \qquad x_0 \in [0,1]^4,
\]
where $N \colon [0,1]^4 \to [0,1]^4$ is a feedforward $\ReLU$ network
of width $4$. In other words, the object of study is a recurrent network that takes no per-step input: its input is its
initial state $x_0$, and its outputs are its states $x_k$. Iterating
one fixed block is the pattern behind  deep equilibrium
models~\citep{bai2019deep} and discretized Neural
ODEs~\citep{chen2018neural,kidger2022neural}. Residual and highway
networks~\citep{he2016deep,srivastava2015highway} follow the same
pattern with untied weights. We ask
what the iterates of a width-$4$ block can compute, and what it costs
to verify their behavior.

\paragraph{The correspondence.}
We prove that iterated width-$4$ $\ReLU$ networks are Turing complete
on \emph{convex} input regions, with iteration count playing the role
of time. 
For every Turing machine (TM) $M$, we construct in
polynomial time a single network $N_M \colon [0,1]^4 \to [0,1]^4$ of
width $4$ and depth $\mathrm{poly}(|M|)$ per iteration. Its iterates
simulate $M$ exactly on encoded tapes, and irreversibly flag every
point of $[0,1]^4$ whose simulation reads an invalid digit (\Cref{thm:main}).
Instantiated on a universal TM $U$, the construction yields a single fixed block $N_U$
that simulates every machine on every input.
The consequence is a hardness ladder along the single axis of the
time bound (\Cref{thm:verif-spectrum}). At fixed width $4$,
bounded-time reachability is $\NP$-complete for a unary bound,
$\NEXP$-complete for a binary bound and undecidable with no bound.
$\NP$-hardness of $\ReLU$ verification has been known
since~\citet{katz2017reluplex}; the ladder shows that 
restricting the width does not restore tractability,
 since width $4$ already suffices
(\Cref{sec:motivation}). The hardness extends at width $6$ to every common smooth
activation  (\Cref{thm:rho}) and, through the
correspondence of~\citet{yifei2024vanilla}, to discretized Neural
ODEs at dimension~$5$ (\Cref{thm:neural-ode}).

\paragraph{The obstruction, in plain terms.}
A TM tape is an infinite binary string, and to feed it to a network we must store it in a single real number. Writing the tape in base
$2$ fails twice over: $0.0\overline{1}_2 = 0.1_2$, so two different
tapes can be the same number, and reading off the first bit is a
discontinuous operation. Writing it in base $4$ using only the digits
$1$ and $3$, a device going back to~\citet{SS91b}, repairs both
defects. The encoding becomes injective, and the two possible leading
digits are separated by a fixed gap, so a continuous piecewise-affine
gadget can read the first digit. The price is that the valid
encodings form a Cantor-like set $\onethreefouradic$ of measure zero,
whereas verification asks a question about a full-dimensional convex
region. The two domains cannot be reconciled by any continuous
preprocessing: any continuous map from a convex set into
$\onethreefouradic$ is constant (\Cref{prop:obstruction}), and
$\ReLU$ networks compute only continuous piecewise-affine maps. Since
continuous encoding is impossible, our network \emph{detects}: a fourth coordinate accumulates, one
base-$4$ digit per iteration, a continuous certificate that
irreversibly flags every input whose simulation reads an invalid digit. This
online certificate is the ingredient that is new at the level of
network construction. It is what makes every verification
reduction below sound on convex regions.
 
\paragraph{Four theorems.}
Formal statements are in~\Cref{sec:def}, proofs in the appendices.

\noindent\textbf{(Th.\ref{thm:main}) Turing completeness of iterated
width-$4$ networks on convex input spaces.} For every TM $M$, a
single polynomial-time-constructible width-$4$ $\ReLU$ network
simulates $M$ when iterated, exactly on encoded inputs, and irreversibly
flags every point of the convex cube $[0,1]^4$ whose simulation reads an invalid digit.

\noindent\textbf{(Th.\ref{thm:verif-spectrum}) Verification
spectrum.} Bounded-time reachability for iterated width-$4$ $\ReLU$
networks is $\NP$-complete with unary time bound, $\NEXP$-complete
with binary time bound and $\RE$-complete (undecidable) with no
bound.

\noindent\textbf{(Th.\ref{thm:rho}) Robust verification at width $6$ for smooth
activations.} For every continuous, non-polynomial,
$C^1$-at-a-point, polynomial-time-computable activation $\rho$, the
$\delta$-robust verification problem (a promise problem with
$\delta/2\delta$ gap) for width-$6$ $\rho$-networks is $\NP$-complete.

\noindent\textbf{(Th.\ref{thm:neural-ode}) Neural ODE consequences at dimension $5$}
(App.\ref{app:neural-ode}). Through the correspondence
of~\citet{yifei2024vanilla} between vanilla feedforward networks and
splitting-method discretizations, the whole spectrum transfers to
the discretized reachability of Neural ODEs at dimension $5$.

\vspace{-0.2cm}
\paragraph{Positioning.}
\Cref{tab:positioning} situates the results; its mode column records
whether a statement concerns a feedforward network or the iterates of
one (a recurrent network).
Siegelmann--Sontag~\cite{LivreSiegelmann,SS91b,SS94} established TM
universality for recurrent networks; in the variant relevant here,
where the input is placed in the initial state, their construction
requires fractal-encoded inputs and does not optimize width. (Their
results with per-step input sequences carry no such restriction, but
answer a different question.) Bounded-width \emph{approximation} by
feedforward networks was developed
by~\cite{hanin2017approximating,park2020minimumwidth,telgarsky2016benefits}.
\citet{MFCS2025} contains two distinct results: a width-$3$ network
simulating one TM step exactly on the fractal encoding set, and a
separate width-$12$ uniform approximation theorem for arbitrary
continuous piecewise-affine maps on convex domains.
\Cref{thm:main} improves neither width. It establishes a different
property: a single width-$4$ network whose iterates are exact on
encoded inputs and certify non-encoded inputs, uniformly over the
whole convex cube $[0,1]^4$ (\Cref{sec:main-sketch-obstruction}
explains why neither prior result yields this). Verification of
$\ReLU$ networks was previously known to be $\NP$-hard via SMT-style
encodings~\cite{katz2017reluplex}; (Th.\ref{thm:verif-spectrum})
lifts this to the full $\NP/\NEXP/\RE$ spectrum along the time bound,
and (Th.\ref{thm:rho}) extends it to smooth activations at width $6$.
Training hardness is known up to
$\exists\R$-completeness~\cite{blum1988training,abrahamsen2021training,froese2022computational}.
The question we study is also the narrow-network analogue of a line
of work on how expressive power grows with the number of sequential
steps: \citet{MerrillSabharwal2024} show that transformers gain
expressive power as the number of chain-of-thought steps grows, and
ours is the same trade-off in a different architecture class, pushed
to its limit. Finally, the Neural ODE consequence
(Th.\ref{thm:neural-ode}) uses the discrete-to-continuous
correspondence~\cite{chen2018neural,yifei2024vanilla}.\footnote{\label{commpanion} Companion
work~\cite{FSTTCS2026} establishes $\NP$-completeness at
widths $3$ and $4$ via fractal preprocessing from 3-SAT, with
$\#\mathrm{P}/\mathrm{PP}$ corollaries on the fractal encoding set.
The TM-simulation, $\NEXP$, undecidability and smooth-activation
contributions of the present paper are disjoint; the only overlap is
\Cref{cor:np-verif-width4} ($\NP$-completeness of width-$4$
verification on convex input), which here arises as a corollary of
\Cref{thm:main} (a strictly stronger primitive whose further
consequences are illustrated by
\Cref{thm:verif-spectrum,thm:rho,thm:neural-ode}
and follow more generally from standard complexity
theory~\cite{Sip97,arora2009}).}
\begin{table}[h]
\vspace{-0.5cm}
\centering\footnotesize
\setlength{\tabcolsep}{3pt}
\renewcommand{\arraystretch}{1.15}
\begin{tabular}{@{}l c c c c c@{}}
\toprule
                & mode & input domain        & guarantee               & width & activation \\
\midrule
\makecell[l]{Siegelmann--Sontag\\ \citeyearpar{SS91b,SS94,LivreSiegelmann}}
                & recurrent
                & fractal $\onethreefouradic$
                & exact TM simulation
                & \makecell{not\\ optimized}
                & \makecell{saturated\\ linear} \\
\makecell[l]{\cite{hanin2017approximating}\\\cite{park2020minimumwidth}}
                & feedforward
                & compact $\R^d$      & uniform approx.    & $d{+}1$ to $d{+}4$
                & $\ReLU$ \\
\makecell[l]{\cite{MFCS2025},\\ main thm.}
                & feedforward
                & convex $[0,1]^d$    & \makecell{uniform approx.\\ of CPAMs}   & $12$   & $\ReLU$ \\
\makecell[l]{\cite{MFCS2025},\\ Thm.~23}
                & iterated
                & fractal $\onethreefouradic$
                & \makecell{exact TM step;\\ no claim off fractal}
                & $3$ & $\ReLU$ \\
\cite{FSTTCS2026}
                & iterated
                & fractal + convex    & $\NP$-hardness via 3-SAT  & $3$, $4$ & $\ReLU$ \\
\textbf{this paper, Th.\ref{thm:main}}
                & iterated
                & convex $[0,1]^4$    & \makecell{\textbf{exact TM simulation}\\ \textbf{+ certificate}}
                & \textbf{$4$} & $\ReLU$ \\
\textbf{this paper, Th.\ref{thm:rho}}
                & both
                & convex $[0,1]^d$    & robust verif.\ $\NP$-c.      & \textbf{$6$}
                & smooth $\rho$ \\
\bottomrule
\end{tabular}
\caption{Positioning. The mode column distinguishes feedforward
statements from statements about the iterates of a fixed block.
(Th.\ref{thm:main}) improves neither width of \citet{MFCS2025}: it
is the first exact TM simulation whose soundness extends to a
full-dimensional convex input region, which is what the $\NEXP$ and
undecidability statements require.}
\label{tab:positioning}

\vspace{-0.5cm}
\end{table}

 \vspace{-0.2cm}
\section{Model, problems, and main theorems}
\label{sec:def}
 
\subsection{Networks, iteration, and verification}
\label{sec:def-model}

\paragraph{The model: iterated networks.}
The object of study in \Cref{thm:main,thm:verif-spectrum} is the
discrete-time system $x_{k+1} = N(x_k)$ obtained by iterating a fixed
feedforward $\ReLU$ network $N \colon [0,1]^4 \to [0,1]^4$ of width
$4$. In other words, it is a weight-tied recurrent network that takes
no per-step input: its input is its initial state and its outputs are
its states. We write $N^{[t]} := N \circ N \circ \cdots \circ N$
for the $t$-fold composition. Which representation a result concerns
is stated with the result: the unary-bound $\NP$-completeness also
holds for feedforward width-$4$ networks, by unrolling
(\Cref{cor:np-verif-width4}), while the $\NEXP$ and $\RE$ statements
concern the iterated representation, whose unrolled depth would be
exponential or unbounded.

We consider neural networks with rational weights, with the
``enhanced neuron'' convention (e.g.~\citet{kidger2020universal}): a neuron
is a map $\psi \circ \rho \circ \phi$ with $\phi, \psi$ affine and
$\rho$ the activation. We write
$\Netnotation{\rho}{n}{m}{k}$ for the class of feedforward networks
with $n$ inputs, $m$ outputs, $\rho$ activation, and bounded width $k$
(i.e. every hidden layer has at most $k$ neurons).

\paragraph{Verification problems.}
Our hardness results concern the following decision problems.
 
\begin{definition}[Reachability]\label{def:reach}
The \emph{reachability} problem takes as input a $\rho$-network
$N \colon [0,1]^d \to [0,1]^d$ and two convex regions $\Init, \Acc \subseteq [0,1]^d$
specified by rational linear inequalities, and asks whether
\(
\exists\,x \in \Init,\ \exists\,t \ge 0,\ N^{[t]}(x) \in \Acc.
\)

Here, the network is presented as a weighted graph, where the encoding of each hidden layer is the adjacency matrix, but every entry is the numerator-denominator-pair describing the rational weight. This formulation follows the standard one in the neural-network verification literature (see e.g. \citealp{katz2017reluplex,wurm1}).
\end{definition}
 
\begin{definition}[Bounded-time reachability]\label{def:bdd-reach}
The \emph{$T$-bounded reachability} problem takes the same input plus a
time bound $T \in \N$ (encoded either in unary or in binary) and asks
whether
\(
\exists\,x \in \Init,\ \exists\,t \le T,\ N^{[t]}(x) \in \Acc.
\)
\end{definition}
 
When $\rho = \ReLU$, both problems admit $\NP$-membership for unary $T$:
by the small-witness property for piecewise-affine systems
(Lemma \ref{lem:np-witness}, classical), it suffices to guess a rational $x$
of polynomial bit complexity and simulate $T$ steps with exact rational
arithmetic. For binary $T$, no such polynomial-time simulation exists;
we recover $\NEXP$-membership in~\Cref{sec:zoo}.

\vspace{-0.2cm}

\subsection{Why reachability, why convex regions, why iterates}
\label{sec:motivation}

Robustness certification is exactly the reachability question we
study: an $\ell_\infty$ ball around a data point $\bar x$ is an
axis-aligned box, hence a convex polytope specified by rational
linear inequalities, and the certification query is literally
$\exists x \in \Init \colon N(x) \in \Acc$, with $\Acc$ the region
where the decision is wrong. This is the VNN-LIB and VNN-COMP
specification format~\citep{brix2023vnncomp}, decided by complete
verifiers such as Reluplex, Marabou and
$\alpha,\beta$-CROWN~\citep{katz2017reluplex,wang2021beta}. Convex
initial regions are therefore imposed by the application. Iteration
matters wherever one
block is applied repeatedly, above all in closed-loop verification
of a neural controller, where the reachable set over $k$ control
steps is exactly $N^{[k]}(\Init)$. The practical reading of our results is that
certification cost scales along depth, or number of steps, and not
along width: at width $4$, relaxing only the time bound climbs
$\PTIME \to \NP \to \NEXP \to$ undecidable, and the remaining route to
tractability is a restriction on the dynamics or on the parameters
(\Cref{sec:discussion}).

\vspace{-0.2cm}

\subsection{Encoding Turing machines}
\label{sec:def-encoding}
\label{sec:enc}

We encode Turing-machine (TM) configurations using base-$4$ digits
$\{1,3\}$, a Cantor-style encoding in the lineage
of~\cite{SS91b,SS94}. Allowing all digits would create ambiguities
like $0.1\overline{3}_4 = 0.2_4$; restricting to digits $1$ and $3$
makes the presentation of a number unique, and resolves the
discontinuity issues inherent to $\ReLU$ iteration. A word $u = u_1\cdots u_n \in \{0,1\}^n$
maps to $\enc_{\{1,3\}}(u) = \sum_{i=1}^{n} (2 u_i + 1)\, 4^{-i} \in [0,1]$.
A configuration $c = ( u, q, v)$ (current state $q$, tape to the
left of the head $u$ (reversed), tape from the head onward $v$)
is encoded as
\[
   \enc_M(c) \;=\; \bigl(\enc_{\{1,3\}}(u),\;
                          \enc_M(q, v),\;
                          0,\; 0\bigr) \;\in\; [0,1]^4,
\]
where $\enc_M(q, v)$ co-encodes the state and the right tape into a
single base-$4$ string with $\enc_M(q)$ as a constant-length prefix
followed by $\enc_{\{1,3\}}(v)$, based on  a packing borrowed
from~\cite{MFCS2025} that saves one coordinate over the naive
encoding $(\enc_{\{1,3\}}(u), \enc_{\{1,3\}}(v), \enc_M(q), \cdot, \cdot)$
which would force width~$5$. The third coordinate is a scratch register
used by the simulator; the fourth is a validity register, both
initially~$0$ (see \Cref{sec:zoo}). We write $Q$ for the state set
of $M$, $\enc_M(Q) \subseteq [0,1]$ for the prefix-encoded set of
states, $\onethreefouradic \subseteq [0,1]$ for the Cantor-like set
of points whose base-$4$ expansion uses only digits $\{1,3\}$,
$\pi_i$ for the $i$-th coordinate projection, $\step_M$ for the map
sending a configuration of $M$ to the next one, and
$c_t(\tau) := \step_M^{\,t}(c_0(\tau))$ for the configuration at
step $t$ on an infinite tape $\tau \in \{0,1\}^\omega$, with
$c_0(\tau)$ the initial configuration.
The \emph{state prefix} of $x_2$ consists of its first
$|q|_4 = \lceil \log_4 |Q| \rceil$ base-$4$ digits, and the
\emph{post-state-prefix} (or tape-suffix) portion of $x_2$ is the
remainder. We say that a digit of $x_1$, or of the tape-suffix of
$x_2$, is \emph{visited during the first $k$ iterations} of the
network of \Cref{thm:main} if it is exposed as the leading digit of
its register at some iteration $j < k$. When $x$ encodes a
configuration, the visited digits are exactly the tape cells read by
$M$'s head.

\subsection{Turing completeness of iterated width-4 networks on convex input spaces}
\label{sec:def-T1}

\begin{theorem}[Turing completeness on convex input spaces]\label{thm:main}
For every deterministic Turing machine $M$ over the binary alphabet,
one can construct in polynomial time a single $\ReLU$ network
$N_M \colon [0,1]^4 \to [0,1]^4$ of width $4$ and depth
$D_M = \mathrm{poly}(|M|)$, with a constant $C_M \in \N$, such that:
\begin{enumerate}\setlength{\itemsep}{0pt}
\item[(i)] (Faithful simulation, all time.) For every $\tau \in \{0,1\}^\omega$
and every $t \ge 0$,
$N_M^{[C_M\, t]}\bigl(\enc_M(c_0(\tau))\bigr) = \enc_M\bigl(\step_M^{\,t}(c_0(\tau))\bigr)$.
(For finite inputs $w$, take $\tau = w \cdot \tau'$ for any suffix
$\tau'$; $M$'s behavior up to time $T$ depends only on the first
$\le T$ symbols.)
\item[(ii)] (Online validity certificate.) The fourth coordinate
$\pi_4(N_M^{[k]}(x))$ is non-decreasing in $k$ and bounded in $[0,1]$.
Moreover, for $x=(x_1,x_2,x_3,x_4)$ with $x_4 = 0$, we have $\pi_4(N_M^{[k]}(x)) = 0$ iff every leading base-$4$ digit of $x_1$ and
of the post-state-prefix portion of $x_2$ visited by $M$'s head during
the first $k$ iterations lies in $\{1, 3\}$, and the state prefix of
$x_2$ stays in $\enc_M(Q)$ along the trajectory. In particular,
$\pi_4(N_M^{[k]}(x)) = 0$ for all $k \ge 0$ whenever $x_1$ and the
$v$-part of $x_2$ lie in $\onethreefouradic$ and the state prefix of
$x_2$ encodes a valid state.
\end{enumerate}
Part~(i) alone holds at width $3$ (\Cref{thm:width3-step}, due
to~\citealp{MFCS2025}).
\end{theorem}

\begin{remark}[Online certificate; single fixed network]
\label{rem:online-cert}\label{rem:single-fixed-network}
Part~(ii) is an \emph{online} certificate: the fourth coordinate is
updated one digit per iteration via a width-$2$ sub-gadget, so $N_M$
is uniformly polynomial in $|M|$, independent of input length. (A
finite $\ReLU$ network cannot have its zero set equal to
$\onethreefouradic$ exactly, hence a check spread over the
iterations.) Taking $M = U$ a universal Turing machine,
$\enc_U(\langle M', w\rangle)$ embeds an arbitrary pair $(M', w)$ into
a single point of $[0,1]^4$, and $N_U$ becomes a single fixed network
simulating any TM $M'$ on any input $w$.
Width $4$ is the price of the pair (i)+(ii), and we claim no lower
bound for the pair.
\end{remark}
The combination of (i) and (ii) is what makes the verification
reductions work: property (i) gives soundness on encoded inputs, and
property (ii) on the rest of the convex region. Plugging in halting
problems of varying time complexity yields the hardness spectrum
below.

\begin{remark}[Convex inputs are the range of a quantifier, not a tolerance]
\label{rem:not-robustness}
\Cref{thm:main} is not a robustness property: part~(i) requires an
exact encoding, and nothing is claimed about inputs near an encoding
beyond part~(ii). The convex region is the range of the existential
quantifier in the verification problems, where it is imposed by the
application (\Cref{sec:motivation}); part~(ii) is what holds on the
rest of that region.
\end{remark}

\begin{remark}[No certificate hypothesis in the complexity results]
\label{rem:certify}
Certifying, in the sense of part~(ii), is a capability of the witness
network, not a hypothesis on a model class:
\Cref{def:reach,def:bdd-reach,thm:verif-spectrum} and
\Cref{cor:np-verif-width4,cor:forall-verif} quantify over \emph{all}
width-$4$ $\ReLU$ networks. \Cref{rem:certify-app} makes the
certified set precise.
\end{remark}
\vspace{-0.3cm}

\subsection{Verification spectrum of iterated width-4 networks}
\label{sec:def-T2}

\begin{theorem}[Verification spectrum at width $4$]\label{thm:verif-spectrum}
For iterated width-$4$ $\ReLU$ networks $N \colon [0,1]^4 \to [0,1]^4$
(the system of \Cref{sec:def-model}), $T$-bounded reachability is:
\vspace{-0.1cm}
\begin{enumerate}\setlength{\itemsep}{-5pt}
\item[(i)] $\NP$-complete when $T$ is encoded in unary;
\item[(ii)] $\NEXP$-complete when $T$ is encoded in binary;
\item[(iii)] $\RE$-complete (hence undecidable) with no time bound.
\end{enumerate}
\vspace{-0.1cm}
The hardness statements hold even when $\Init$ and $\Acc$ are axis-aligned
hyperrectangles with rational endpoints.
\end{theorem} 

\begin{corollary}[Verification of feedforward width-$4$ $\ReLU$ networks]
\label{cor:np-verif-width4}
Verification of \emph{feedforward} width-$4$ $\ReLU$ networks on convex
$\Init, \Acc \subseteq [0,1]^4$ (deciding whether
$\exists x \in \Init : N(x) \in \Acc$, a single application of $N$) is $\NP$-complete.
\end{corollary}

\subsection{Robust verification at width 6 for smooth activations}
\label{sec:def-T3}
 
\paragraph{Why ``robust''.}
Exact verification of $\rho$-networks for smooth $\rho$ such as $\tanh$
quickly leaves classical complexity classes: the relevant decision
problems involve transcendental constraints whose decidability rests
on contested
hypotheses~\citep{macintyre1996decidability,wurm1,wurm2}. The standard
remedy is robust verification, which allows a fixed tolerance with a
promise gap.
 
\begin{definition}[$\delta$-robust verification]\label{def:robust-verif}
Given a $\rho$-network $N \colon [0,1]^d \to [0,1]^d$, convex regions
$\Init, \Acc \subseteq [0,1]^d$, and a rational $\delta > 0$, the
$\delta$-\emph{robust verification problem} is the \textbf{promise} problem:
\vspace{-0.1cm}
\begin{itemize}\setlength{\itemsep}{-3pt}
\item Output \textsc{Yes} if $\exists\,x \in \Init$ with $N(x) \in \Acc^{+\delta}$;
\item Output \textsc{No} if $\forall\,x \in \Init$, $N(x) \notin \Acc^{+2\delta}$;
\item Promise: the input falls into one of these two cases.
\end{itemize}
Here $\Acc^{+\delta} = \{y \in \R^d : \dist(y, \Acc) \le \delta\}$ and
$\dist$ is Euclidean distance. The $T$-bounded variant considers
$N^{[t]}(x)$ for $t \le T$.
\end{definition}
 
\begin{theorem}[Robust verification at width $6$]\label{thm:rho}
Let $\rho \colon \R \to \R$ satisfy:
\begin{enumerate}\setlength{\itemsep}{-3pt}
\item[(H1)] $\rho$ is continuous and non-polynomial;
\item[(H2)] $\rho$ is $C^1$ at some point with non-zero derivative;
\item[(H3)] $\rho$ is polynomial-time computable, and admits a
polynomial-time-computable Lipschitz constant on every bounded interval.
\end{enumerate}
Then $\delta$-robust verification and $\delta$-robust $T$-bounded
reachability (with $T$ unary) for $\rho$-networks of width $6$ are
$\NP$-complete\footnote{Both are promise problems with a $\delta$/$2\delta$
gap; ``$\NP$-complete'' is shorthand for ``$\NP$-hard for the YES
instances of the promise and admitting an $\NP$ verifier on the
promise,'' i.e. formally $\promiseNP$-complete in the sense of
\cite{goldreich2006promise}.}.
\end{theorem}

\paragraph{Coverage of (H1)--(H3).}
(H1)--(H2) is the relaxation of differentiability
of~\citet{kidger2020universal}; with (H3) they cover $\ReLU$, leaky
$\ReLU$, $\tanh$, sigmoid, GELU, Swish, ELU, softplus, $\arctan$,
the Gaussian and every analytic non-polynomial activation.
Constructive $\rho \to \ReLU$ bit-complexity bounds are provided
by~\citet{zhang2024deep}; the remaining cases follow from interval
evaluation under (H3).
 
 \vspace{-0.2cm}

\section{Proof sketch of Theorem~\ref{thm:main}}
\label{sec:main-proof-sketch}

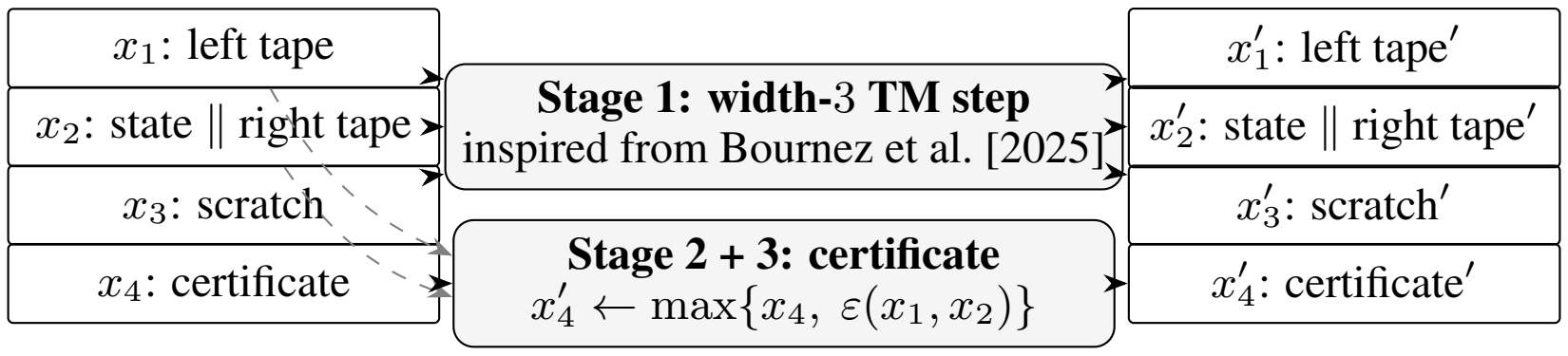
\begin{figure}[t]
\centering
\begin{tikzpicture}[scale=0.78,
                    reg/.style={draw, rounded corners=1pt,
                                minimum width=3.0cm, minimum height=0.55cm,
                                font=\small},
                    block/.style={draw, fill=gray!8, rounded corners,
                                  minimum width=4.6cm, minimum height=0.85cm,
                                  align=center, font=\small}]
\node[reg] (r1) at (0, 3.0) {$x_1$: left tape};
\node[reg] (r2) at (0, 2.3) {$x_2$: state $\Vert$ right tape};
\node[reg] (r3) at (0, 1.6) {$x_3$: scratch};
\node[reg] (r4) at (0, 0.9) {$x_4$: certificate};
\node[block] (sim) at (5, 2.3)
  {\textbf{Stage 1: width-$3$ TM step}\\
inspired from~\cite{MFCS2025}};
\node[block] (cert) at (5, 0.9)
  {\textbf{Stage 2 + 3: certificate}\\
   $x_4' \leftarrow \max\{x_4,\ \Cert(x_1,x_2)\}$};
\node[reg] (o1) at (10, 3.0) {$x_1'$: left tape$'$};
\node[reg] (o2) at (10, 2.3) {$x_2'$: state $\Vert$ right tape$'$};
\node[reg] (o3) at (10, 1.6) {$x_3'$: scratch$'$};
\node[reg] (o4) at (10, 0.9) {$x_4'$: certificate$'$};
\foreach \i in {1,2,3} {\draw[-{Stealth}] (r\i) -- (sim);}
\foreach \i in {1,2,3} {\draw[-{Stealth}] (sim) -- (o\i);}
\foreach \i in {1,2} {\draw[-{Stealth}, dashed, gray] (r\i) to[bend right=18] (cert);}
\draw[-{Stealth}] (r4) -- (cert);
\draw[-{Stealth}] (cert) -- (o4);
\end{tikzpicture}
\caption{One iteration of the width-$4$ network of \Cref{thm:main}:
$x_1$ holds the left tape, $x_2$ co-encodes the state and the right
tape, $x_3$ is a scratch register, and $x_4$ accumulates the
certificate $\Cert$ via the max update, with $\Cert(x_1, x_2) = 0$
iff the leading digits and the state prefix are valid
(Remark~\ref{rem:checking-applies-twice}).
}
\vspace{-0.5cm}
\label{fig:width4-schematic}
\end{figure}

The proof of \Cref{thm:main} has three modular components, illustrated
in \Cref{fig:width4-schematic}: a width-$3$ subnetwork that simulates
one Turing-machine step on the $\{1,3\}$-base-$4$ encoding domain
(\emph{Stage~1}, at width $3$, based on ideas  
from~\cite{MFCS2025}); a width-$2$ subnetwork that produces a
continuous encoding-validity certificate (\emph{Stage~2}); and a combinator that propagates the
certificate across iterations (\emph{Stage~3}). The full network has
width~$4$. We first identify the obstruction, then describe each
stage; complete proofs are in
\Cref{app:width3,app:encoding,app:main-proof}.

\subsection{The continuity obstruction, and why exactness is forced}
\label{sec:main-sketch-obstruction}

\paragraph{Why exactness, in one sentence.}
Each simulated tape shift multiplies a coordinate by $4$
(\Cref{lem:shift}), so a per-step error $\epsilon$ becomes about
$4^k \epsilon$ after $k$ iterations: an approximation accurate for
$k$ steps needs precision $4^{-k}$, hence a network whose size grows
with the time horizon. \Cref{thm:verif-spectrum}(ii)--(iii) quantify
over exponential and unbounded $k$ with one fixed finite network, so
no approximation-based construction can reach them.

\paragraph{The obstruction.}
The existing $\ReLU$-based simulations operate on the fractal set
$\onethreefouradic$ (totally disconnected, Lebesgue measure $0$),
whereas verification asks for a trajectory from a \emph{convex}
initial region $\Init \subseteq [0,1]^4$. The two domains are
incompatible in a strong sense.

\begin{proposition}[No continuous encoding]\label{prop:obstruction}
Every continuous map from a convex subset of $\R^n$ into
$\onethreefouradic$ is constant.
\end{proposition}

\begin{proof}
A convex set is connected, the continuous image of a connected set is
connected, and $\onethreefouradic$ is totally disconnected, so the
image is a single point.
\end{proof}

$\ReLU$ networks realize only continuous piecewise-affine
maps~\cite{arora2016understanding}, so no subnetwork can map $\Init$
onto valid encodings, and none can gate the simulation on the
discontinuous predicate ``is $x$ a valid encoding?''.
\cite{MFCS2025} circumvent this by approximating the validity
predicate uniformly, at width $12$ and at the cost of exact
iteration; the companion work~\cite{FSTTCS2026} restricts inputs by
3-SAT preprocessing, which yields $\NP$-hardness but no TM
simulation. We instead realize a \emph{continuous} certificate
$\Cert(x)$ that vanishes on valid encodings, is positive elsewhere
and is non-decreasing along trajectories: any input which was never a
valid encoding is flagged and stays flagged. The certificate occupies
one extra coordinate, hence width~$4$.

\paragraph{Stage 1: width-$3$ TM step on encoded inputs.}
\label{sec:main-sketch-stage1}
TM simulation at some fixed width has been known
since~\cite{SS91b}; \cite{MFCS2025} reach width $3$, which we make
fully explicit in \Cref{app:width3} and use as a black box.

\begin{theorem}[Width-$3$ step simulation, inspired from {\cite{MFCS2025}}]
\label{thm:width3-step}
For every Turing machine $M$ over $\{0,1\}$, there exists a $\ReLU$
network $g_M \in \Netnotation{\ReLU}{3}{3}{3}$ of depth $\mathrm{poly}(|M|)$
such that, for every valid encoded configuration
$\enc_M(c) \in \onethreefouradic \times {\onethreefouradic}_{Q} \times \{0\}$,
\(
   g_M\bigl(\enc_M(c)\bigr) = \enc_M\bigl(\step_M(c)\bigr),
\)
where ${\onethreefouradic}_{Q} \subseteq [0,1]$ denotes the set of
co-encodings of a state prefix in $\enc_M(Q)$ followed by an
$\onethreefouradic$-tape suffix, and the third coordinate is a scratch
register reset to $0$ between steps.
\end{theorem}

\Cref{thm:width3-step} treats the encoding domain as a set of exact
rational points: each TM transition is an explicit affine map on the
relevant cylinder of $\onethreefouradic \times {\onethreefouradic}_{Q}$,
and the gaps between cylinders allow these maps to be assembled at
width $3$. Outside the encoded domain, $g_M$ has no specified
behavior, which is what forces the certificate: it is what lets us
talk about verification on convex inputs rather than on the fractal
encoding set.

\vspace{-0.2cm}

\paragraph{Stage 2: continuous encoding-validity certificate.}
\label{sec:main-sketch-stage2}
The new ingredient is a per-step error gadget
$\Err \colon [0,4] \to [0,1]$, continuous and piecewise affine,
which vanishes exactly on $[1,2]\cup[3,4]$ and is positive elsewhere. It can
be implemented as a sum of two tents by the nested ReLU expression
\\
\centerline{
$
   \Err(x)
   = \ReLU\!\Bigl(
       3 - x - \ReLU\!\bigl(
                  5 - 2x - \ReLU\!\bigl(
                              2 - x + \ReLU(1-x)
                            \bigr)
                \bigr)
     \Bigr).
$}
Applied to $4x_j$, $\Err$ vanishes iff the leading base-$4$ digit of
$x_j$ lies in $\{1, 3\}$ (scheduling inside the width-$4$ network:
\Cref{app:encoding}).

\begin{proposition}[Per-step encoding-validity gadget]
\label{prop:certificate}
The function $\Err_4(x):=\Err(4\cdot x)$ is continuous, non-negative, computable
in constant depth inside the width-$4$ register network of
\Cref{thm:main}, and vanishes precisely on the leading-digit
$\{1,3\}$-cylinders of $[0,1]$.
\end{proposition}

\paragraph{Stage 3: combining the stages, and proof of \Cref{thm:main}.}
\label{sec:main-sketch-stage3}

The full width-$4$ network $N_M$ assembles, at each iteration, a
\emph{per-step certificate update} followed by Stage~1's TM step:
\begin{enumerate}\setlength{\itemsep}{-3pt}
\item Apply the per-step error gadget $\Err$ to $4 x_1$ and to the
leading digits of the tape-suffix portion of $x_2$
(\Cref{prop:certificate}; the state-prefix portion of $x_2$ is checked
by the same gadget composed with a constant-depth slicing,
\Cref{app:encoding}). Update
$x_4' \leftarrow \max\{x_4,\,\Cert(x_1, x_2)\}$, where
$\Cert(x_1, x_2)$ is the maximum of the digit-error and
state-distance terms (Remark~\ref{rem:checking-applies-twice}).
The maximum is computed as $a + \ReLU(b - a)$ at width $2$.
\item Apply Stage~1, the width-$3$ step simulation $g_M$, to
$(x_1, x_2, x_3)$ to obtain $(x_1', x_2', x_3')$. The fourth register
$x_4'$ is propagated unchanged through Stage~1 by an identity wire.
\end{enumerate}
Each iteration of $N_M$ has constant depth on top of $g_M$; thus the
per-iteration depth of $N_M$ is $D_M = \mathrm{poly}(|M|)$, independent
of any input length or time horizon.

\paragraph{Scheduling and width-$4$ bound.} Each step is a sequence
of register-layer updates (\Cref{def:relu-register-net}), each
modifying one of $(x_1, \dots, x_4)$: the error gadgets are
width-$2$ and sequentialized so that $x_4$ serves as accumulator
while $x_1, x_2$ are read-only and $x_3$ holds scratch data; Stage~1
then updates $x_1, x_2, x_3$ and preserves $x_4$ through an identity
wire (register form of register width $4$,
Remark~\ref{rem:nm-register}).

\paragraph{Proof of \Cref{thm:main}.}
$\Cert$ is non-negative and bounded by $1$; the $\max$ update keeps
$x_4(k) := \pi_4(N_M^{[k]}(x))$ non-decreasing and in $[0,1]$. If
$x_1$ and the tape-suffix of $x_2$ lie in $\onethreefouradic$, and the
state-prefix of $x_2$ is in $\enc_M(Q)$, every leading digit visited
stays in $\{1,3\}$ and the state stays in $\enc_M(Q)$ along the
trajectory (since $g_M$ preserves valid encodings by
\Cref{thm:width3-step}); $\Cert$ vanishes at every iteration, $x_4 = 0$
throughout, and $N_M^{[t]}(x) = \enc_M(c_t)$ by induction, proving
(i). Conversely, when a visited leading digit falls outside
$[1,2] \cup [3,4]$ or the state-prefix leaves $\enc_M(Q)$, the
per-step error is positive at that iteration; $x_4$ becomes positive
and stays so. Stage~2 checks \emph{before} Stage~1 reads, so the
simulator never operates on a forbidden digit while $x_4 = 0$. \qed\
(A fully worked example is given in \Cref{app:worked-example}.)

\vspace{-0.2cm}

\section{Verification spectrum of iterated width-4 networks}
\label{sec:zoo}

\vspace{-0.2cm}

\subsection{Proof of Theorem~\ref{thm:verif-spectrum}}
\label{sec:zoo-proof}

\paragraph{Reductions: a uniform recipe.}
For each part of \Cref{thm:verif-spectrum}, we reduce a classical
TM complexity problem $\Pi$ to width-$4$ reachability. The reduction
is identical across the three parts, so we describe it once. Let
$\Pi$ ask whether some Turing machine $M$ accepts some input within
some (problem-dependent) time bound.
By \Cref{thm:main}, applied to $M$, we obtain a width-$4$ $\ReLU$
network $N_M$ with the simulation and certificate properties. Recall
that the second coordinate $x_2$ co-encodes the state (constant-length
prefix $\enc_M(q)$ of length $|q|_4 = \lceil \log_4 |Q| \rceil$) and
the right tape (\Cref{sec:def}). Define
\[
   \Init \;=\; \bigl\{ x \in [0,1]^4 \;:\;
     \enc_M(q_0) \le 4^{|q|_4} x_2
                          \le \enc_M(q_0) + 1,\;
     x_3 = 0,\; x_4 = 0 \bigr\},
\]
i.e. $x_2$ has the prefix $\enc_M(q_0)$ of the initial state, the
remaining digits of $x_2$ encode an arbitrary right tape, $x_1$ is
unrestricted in $[0,1]$ (left tape), and the scratch and validity
registers are pinned to $0$. Define
\(
   \Acc \;=\; \bigl\{ x \in [0,1]^4 \;:\;
     \enc_M(q_F) \le 4^{|q|_4} x_2 \le \enc_M(q_F) + 1,\;
     x_4 \le 0 \bigr\},
\)
i.e. the state-prefix of $x_2$ encodes the final state $q_F$. Both
$\Init$ and $\Acc$ are convex and specified by rational linear
inequalities (axis-aligned in $x_2, x_3, x_4$). The reduction maps
$(M, T)$ (or $(M)$ in the unbounded case) to $(N_M, \Init, \Acc, T)$
(or $(N_M, \Init, \Acc)$).
Correctness is by \Cref{thm:main}: a trajectory enters $\Acc$ iff
$x_4 = 0$ throughout, i.e. every step is locally valid in the sense
of Lemma~\ref{lem:invariant}, and the state-prefix of $x_2$ encodes
$q_F$ at some iteration $\le T$. Equivalently, the trajectory enters $\Acc$ iff $M$
accepts on some infinite binary tape $\tau \in \{0,1\}^\omega$ within
the time bound, which matches the $\NP/\NEXP/\RE$ hardness sources by
padding: acceptance on some infinite tape in $T$ steps reduces to
acceptance on some length-$T$ prefix in $T$ steps (see
Lemma \ref{lem:tm-complexity}).

It remains only to choose $\Pi$ for each part and verify the upper bound.

\vspace{-0.2cm}
\paragraph{Part (i): unary $T$.}
Take $\Pi$ to be ``does $M$ accept some $w \in \{0,1\}^n$ within $T$
steps?'' with $T$ unary, which is $\NP$-complete by Cook--Levin
applied to a TM with the standard polynomial slowdown
\cite{arora2009}. Hardness follows from the recipe above. NP-membership
follows from Lemma~\ref{lem:np-witness}: a non-deterministic verifier guesses
a rational $x \in \Init$ of polynomial bit complexity and simulates
$N_M$ exactly for $T \cdot C_M$ steps. With $T$ unary, total time
is $\mathrm{poly}(|N_M|, T)$.

\vspace{-0.3cm}
\paragraph{Part (ii): binary $T$.}
Take $\Pi$ to be \textsc{Succinct-TM-Acceptance}: ``does $M$ accept some
$w$ of length at most $n$ in at most $T$ steps?'' with $n, T$ binary.
This problem is $\NEXP$-complete by a standard padding argument from
\textsc{Succinct-Sat}~\cite{arora2009}.
Hardness follows from the recipe above with $T \cdot C_M$ binary as the
new time bound.
For membership: a non-deterministic verifier guesses $x \in \Init$ and
$t \le T \cdot C_M$ in time polynomial in the input size; simulation of
$N_M$ for $t$ iterations takes time $\mathrm{poly}(t, |N_M|) = 2^{\mathrm{poly}(\text{input})}$
because $t$ may be exponential in the binary encoding $|T|$. The
overall verifier runs in non-deterministic exponential time, hence
$\NEXP$-membership.

\vspace{-0.3cm}
\paragraph{Part (iii): no time bound.}
Take $\Pi$ to be the halting problem on the empty input: ``does $M$
halt on the empty word?'', which is $\RE$-complete. The recipe gives
$N_M$ together with the singleton $\Init = \{\enc_M(c_0(\varepsilon))\}$
(equivalent to a degenerate hyperrectangle with rational endpoints).
A trajectory from this single point reaches $\Acc$ iff $M$ halts on the
empty word. Membership in $\RE$ is a standard semi-decision procedure
(simulate $N_M$ until the trajectory enters $\Acc$, accept if so).
\qed

\noindent\textit{A note on the $\NEXP$ witness.} The witness in
Part (ii) encodes a TM input of length up to $n$ as the point
$x \in \Init$; Remark~\ref{rem:exptime-nexp} explains why this lands
on $\NEXP$ rather than $\EXPTIME$.

\vspace{-0.2cm}
\subsection{Two corollaries used in the sequel}
\label{sec:zoo-corollaries}

Two consequences follow at no additional cost. By complementation, the
dual problem ``does $N(x) \in \Acc$ hold for every $x \in \Init$?'' is
$\coNP$-complete (this is what complete verifiers like Marabou and
$\alpha,\beta$-CROWN~\cite{katz2017reluplex,wang2021beta} certify), and
by a gap construction using the certificate of \Cref{thm:main},
approximating $\min_{x \in \Init,\,t \le T} N^{[t]}(x)$ within additive
error $\eta < 1/4$ inherits the full spectrum.

\begin{corollary}[Universal verification and approximation hardness]
\label{cor:forall-verif}\label{cor:approx}
For width-$4$ $\ReLU$ networks: \textbf{(i)} universal verification is
$\coNP$-complete; \textbf{(ii)} approximating worst-case output within
$\eta$ is $\NP$-hard (unary $T$), $\NEXP$-hard (binary $T$), and
undecidable (no $T$).
\end{corollary}

\section{Robust verification at width 6 for smooth activations}
\label{sec:rho}

\Cref{thm:verif-spectrum} addresses $\ReLU$. We extend the hardness
to every common smooth activation at the cost of two additional
units of width, in the robust framework of \Cref{sec:def-T3}: exact
verification for smooth $\rho$ leaves classical
complexity~\citep{macintyre1996decidability,wurm2,wurm2023complexity},
while the $\delta/2\delta$ promise gap~\citep{gao2012delta} absorbs
the $\rho \to \ReLU$ approximation error and matches the
finite-precision regime of numerical verifiers. \Cref{thm:rho} shows
this relaxation does not help.
\label{sec:rho-why}

\subsection{$\ReLU$-register networks and \texorpdfstring{$\rho$}{rho}-emulation}
\label{sec:rho-register}

The reduction from $\ReLU$ width $4$ to $\rho$ width $6$ is most
cleanly stated in a register normal form, in which each hidden layer
updates a single coordinate. Every $\ReLU$ network of width $w$ can be
expressed in this form at a polynomial cost in depth.

\begin{definition}[$\ReLU$-register network]
\label{def:relu-register-net}
A $\ReLU$-\emph{register network} of register width $k \ge 1$ and
depth $L$ is a composition
\(
   F = A_{\mathrm{out}} \circ R_L \circ \cdots \circ R_1 \circ A_{\mathrm{in}},
\)
where $A_{\mathrm{in}} \colon \R^d \to \R^k$ and
$A_{\mathrm{out}} \colon \R^k \to \R^d$ are affine, and each layer
$R_\ell \colon \R^k \to \R^k$ updates a single coordinate
$j_\ell \in \{1, \dots, k\}$ via
$(R_\ell(x))_{j_\ell} = \ReLU(a_\ell \cdot x + b_\ell)$, leaving the
others unchanged.
\end{definition}

We work directly with this register form: it captures all $\ReLU$
networks (any standard $\ReLU$ network of width $w$ and depth $L$ 
embeds into a register network of register width $2w$ and depth 
$O(wL)$; for our purposes, we only need that $N_M$ is natively in
register form of register width $4$, see Remark~\ref{rem:nm-register} in
\Cref{app:register}). The technical core of the section is the
following emulation result, adapted from~\citet{kidger2020universal}.

\begin{lemma}[Per-layer $\rho$-emulation at width $k+2$]
\label{lem:rho-emulation}
Let $\rho \colon \R \to \R$ satisfy hypotheses (H1)-(H3) of
\Cref{thm:rho}. For every $\ReLU$-register layer $R \colon \R^k \to \R^k$,
every compact $K \subseteq \R^k$, and every $\eta > 0$, there exists a
$\rho$-network $\widetilde R \in \Netnotation{\rho}{k}{k}{k+2}$ of
depth $O(1)$ such that
\(
   \sup_{x \in K} \|\widetilde R(x) - R(x)\|_\infty \;\le\; \eta.
\)
The construction time is polynomial in the description size of $R$ and
in $\log(1/\eta)$.
\end{lemma}

\paragraph{Mechanism of the +2 overhead.}
The width $k+2$ decomposes into $k$ \emph{identity wires} (one per
register coordinate, each implemented by one enhanced $\rho$-neuron
via the local-identity approximation
of~\citet[Lemma~4.1]{kidger2020universal}, see Proposition \ref{kidger41}) plus a
$\rho$-subnetwork of width $2$ that approximates the univariate
non-linearity $\ReLU(a \cdot x + b)$ on a bounded range
(\citet[Proposition~4.9]{kidger2020universal}). The full proof of
Lemma \ref{lem:rho-emulation} is in \Cref{app:rho-emulation}.

Composing per-layer emulators with a standard telescoping argument
yields a network-level approximation result.

\begin{lemma}[Network-level $\rho$-emulation]
\label{lem:rho-emulation-network}
Let $F \colon \R^d \to \R^d$ be a $\ReLU$-register network of register
width $k$ and depth $L$, and let $K \subseteq \R^d$ be compact. For
every $\varepsilon > 0$, there exists a $\rho$-network
$\widetilde F \in \Netnotation{\rho}{d}{d}{k+2}$ of depth $O(L)$, with
$\sup_{x \in K} \|\widetilde F(x) - F(x)\|_\infty \le \varepsilon$, and
constructible in time $\mathrm{poly}(|F|, \log(1/\varepsilon))$.
\end{lemma}

\subsection{Proof of \texorpdfstring{\Cref{thm:rho}}{Theorem 3}}
\label{sec:rho-proof}

We prove $\NP$-membership and $\NP$-hardness of $\delta$-robust
verification for $\rho$-networks of width $6$. The bounded-time variant
(unary $T$) follows by the same argument with depth scaled by $T$.
Throughout this section and \Cref{app:robust-transfer}, we write $R$
for the acceptance region $\Acc$ of \Cref{def:robust-verif}.

\paragraph{$\NP$-membership.}
By Lemma \ref{lem:np-witness} (applied to the $\ReLU$-equivalent given by
Lemma \ref{lem:rho-emulation}), a non-deterministic verifier can guess a
rational $x \in \Init$ of polynomial bit complexity. It then computes
$\widetilde y \approx N(x)$ to precision $\delta/4$ using polynomial-time
evaluation of $\rho$ and exact rational arithmetic for affine layers.
By (H3), each $\rho$ evaluation costs $\mathrm{poly}(\log(1/\delta))$ time and
the polynomial-time-computable Lipschitz constant
$L_\rho = \mathrm{poly}(\text{input})$ on the relevant compact yields total
iteration error $\le T \cdot L_\rho^T \cdot \varepsilon_{\mathrm{step}}$.
Choosing $\varepsilon_{\mathrm{step}} = \delta/(4 T L_\rho^T)$ requires
$\log(1/\varepsilon_{\mathrm{step}}) = \mathrm{poly}(\text{input})$ bits of precision per step.
The verifier returns
\textsc{Yes} iff $\widetilde y \in R^{+3\delta/2}$.
On promise YES, $\dist(\widetilde y, R) \le \delta + \delta/4 = 5\delta/4 \le 3\delta/2$
and the verifier accepts; on promise NO,
$\dist(\widetilde y, R) > 2\delta - \delta/4 = 7\delta/4 > 3\delta/2$
and the verifier rejects.

\paragraph{$\NP$-hardness via robust reduction from $\ReLU$.}
Corollary \ref{cor:approx} (or directly the proof of \Cref{thm:verif-spectrum})
establishes that the $\delta_0$-robust verification problem is
$\NP$-hard for $\ReLU$ networks at width $4$ for some fixed
$\delta_0 > 0$ (concretely, the construction yields a separation of
$\Omega(1)$ between $N(x) \in R$ and $N(x) \notin R^{+1}$ on the
relevant compact). We reduce this to robust $\rho$ verification at
width $6$.

Given a width-$4$ $\ReLU$ instance $(N, \Init, R, \delta)$, 
since $N_M$ is natively in register form of register width $4$ 
(Remark~\ref{rem:nm-register}), apply 
Lemma \ref{lem:rho-emulation-network} on the compact 
$K = [0,1]^4$ with precision $\varepsilon = \delta / 4$ to obtain a
width-$6$ $\rho$-network
$\widetilde N$ such that $\sup_K \|\widetilde N - N\|_\infty \le \delta/4$.
Output the $\rho$-instance $(\widetilde N, \Init, R^{+\delta}, \delta/4)$. \\
\emph{YES preservation.} If the original instance is YES, then
$\exists x \in \Init$ with $N(x) \in R^{+\delta}$. Then
$\dist(\widetilde N(x), R) \le \delta + \delta/4 = 5\delta/4$, so
$\widetilde N(x) \in (R^{+\delta})^{+\delta/4}$. The output instance is
YES. \\
\emph{NO preservation.} If the original is NO, then $\forall x \in \Init$,
$\dist(N(x), R) > 2\delta$, so
$\dist(\widetilde N(x), R) > 2\delta - \delta/4 = 7\delta/4$. We need
$\widetilde N(x) \notin (R^{+\delta})^{+2(\delta/4)} = R^{+3\delta/2}$,
which holds since $7\delta/4 > 3\delta/2$. The output instance is NO.

The reduction is polynomial-time (Lemma~\ref{lem:rho-emulation-network} 
is polynomial-time constructible by 
Lemma~\ref{lem:rho-emulation-network-restated}); promise is
preserved; YES/NO map correctly. Hence $\delta/4$-robust verification
of $\rho$-networks at width $6$ is $\NP$-hard. $\NP$-membership above
gives completeness. \qed

\vspace{-0.2cm}

\paragraph{Two complements.}
The argument needs $T$ unary: with $T$ binary, the precision required
is doubly exponential in the input, and a $\NEXP$-style statement at
binary $T$ is open. The approximation hardness of
Corollary~\ref{cor:approx} transfers to width-$6$ $\rho$-networks by
the same reduction (\Cref{sec:rho-binary}).

\vspace{-0.3cm}

\vspace{-0.3cm}
\section{Discussion and open problems}
\label{sec:discussion}

\paragraph{What this paper establishes.}
The iterates of a single fixed $\ReLU$ network of width $4$
simulate an arbitrary Turing machine, exactly on encoded inputs and
with a certificate on the rest of the convex cube (\Cref{thm:main});
their verification spans the hierarchy from $\NP$ to undecidable
along the time-bound axis (\Cref{thm:verif-spectrum}); the hardness
extends to all standard smooth activations at width~$6$ in the robust
framework (\Cref{thm:rho}) and, through~\citet{yifei2024vanilla}, to
Neural ODEs at dimension~$5$ (\Cref{thm:neural-ode}). The technical
novelty over classical Turing-simulation theorems for narrow recurrent
networks is a continuous encoding-validity certificate that resolves a
continuity obstruction at the cost of one extra unit of width.

\paragraph{Practical implications.}
Depth, not width, is the axis along which verification cost scales
super-polynomially (\Cref{sec:motivation}). This is consistent with the
empirical observation that depth dominates the cost of complete
verifiers~\citep{katz2017reluplex,wang2021beta}, although our
worst-case analysis does not by itself explain solver behavior on
trained networks.

\paragraph{Open problems.}
Four questions remain.
\textbf{(i) Width 2.} Width 1 admits polynomial-time verification,
widths 3 and 4 admit $\NP$-completeness via fractal
preprocessing~\cite{FSTTCS2026}, and width 4 obeys the full spectrum
of \Cref{thm:verif-spectrum}; width 2 is open.
\textbf{(ii) Continuous-time simulation.} \Cref{thm:neural-ode} holds
under splitting-method discretization; whether a Lipschitz
piecewise-affine ODE admits TM simulation by its continuous flow,
at any dimension, is open (\Cref{sec:neural-ode-open}).
\textbf{(iii) Trained and constrained networks.} Our construction is
essentially expanding: each tape shift multiplies by $4$. First,
contractivity: non-expansive iterates (spectral norm at most $1$, as
enforced in deep equilibrium models~\citep{bai2019deep} and
normalizing flows) cannot create digit resolution, and we believe
that bounded-time reachability leaves the $\NEXP/\RE$ regime under
such a constraint. Second, parameterized complexity in the number of
unstable neurons, the quantity that governs complete solvers in
practice, rather than in width or depth.
\textbf{(iv) Beyond piecewise-affine activations.} (H1)--(H3) cover
the standard smooth activations but not softmax or attention.
Extending to Transformer-style architectures is natural.

\begin{ack}
O. Bournez and J. Cohen acknowledge support from PGMO, and O. Bournez
from ANR project OTC ANR-24-CE48-0335. A. Wurm acknowledges partial
support from the BTU Graduate Research School. The authors declare no
competing interests.
\end{ack}
 
 \newpage

\bibliographystyle{plainnat}
\bibliography{reference-biblio,Johanne,camera-ready-additions}
 
\newpage
\appendix
\crefalias{section}{appendix}
\crefalias{subsection}{subappendix}
\crefalias{subsubsection}{subsubappendix}

\section{Notation and preliminaries}
\label{app:prelim}

This appendix collects formal definitions used throughout the paper. The
neural-network model itself is defined in~\Cref{sec:def-model};
here we record additional conventions and the verification problems
considered.

\paragraph{Rational networks.}
Throughout, we restrict attention to \emph{rational} neural networks and
inputs: all parameters (weights, biases, activation slopes and
hyperplane coefficients of $\Init$ and $R$) are rational numbers.
When convenient (here, in \Cref{sec:rho} and in
\Cref{app:robust-transfer}) we write $R$ for the acceptance region
$\Acc$.

\paragraph{Verification and reachability problems.}
Fix an activation $\rho \colon \R \to \R$.

\begin{definition}[$\rho$-Verification]
\label{def:verification}
The \emph{$\rho$-Verification} problem is the following decision problem.
\textbf{Given:} a neural network
$\NN \colon [0,1]^d \to [0,1]^d \in \Netnotatione{\rho}{d}{d}$ and two
convex regions $\Init, \Acc \subseteq [0,1]^d$ specified by linear inequalities.
\textbf{Decide:} is there $x \in \Init$ such that $\NN(x) \in \Acc$?
\end{definition}

\begin{definition}[$\rho$-Bounded-Time Recurrent Reachability]
\label{def:bounded-time-reach}
The \emph{$\rho$-Bounded-Time Recurrent Reachability} problem is the
decision problem.
\textbf{Given:} a neural network
$\NN \colon [0,1]^d \to [0,1]^d \in \Netnotatione{\rho}{d}{d}$, two convex
regions $\Init, \Acc \subseteq [0,1]^d$, and a time bound $T \in \N$ encoded in
unary (resp. binary, or unbounded).
\textbf{Decide:} is there $x \in \Init$ and $t \le T$ with
$\NN^{[t]}(x) \in \Acc$?
\end{definition}

The robust variants ($\delta$-robust verification, $\delta$-robust
bounded-time reachability) are defined in \Cref{def:robust-verif}.

\section{Width-3 simulation of Turing machines (after \cite{MFCS2025})}
\label{app:width3}

This appendix recalls the building block of our main theorem: a
width-$3$ $\ReLU$ network simulating one step of a Turing machine
on a base-$4$ encoding with digits in $\{1,3\}$. The result is
adapted from~\cite{MFCS2025}.

\subsection{Turing machine conventions}

We use the following conventions throughout. A \emph{deterministic
Turing machine} is a tuple
$ M=(Q,\Gamma,\delta,q_0,q_F)$,
where $Q$ is a finite set of states; $\Gamma=\{1,3,B\}$ is the tape
alphabet, with $B$ denoting the distinguished blank symbol;
$ \delta \colon Q\times \Gamma \to Q\times \Gamma\times \{L,R\}$
is the transition function, where $L$ and $R$ denote a one-cell move of
the head to the left and to the right, respectively; $q_0\in Q$ is the
initial state; and $q_F\in Q$ is the unique accepting state. The choice of digits $\{1,3\}$ for the binary input alphabet, instead
of $\{0,1\}$, is consistent with the encoding of~\Cref{sec:enc} and
avoids a separate treatment of the digit $0$. Thus, a binary input
$w\in\{0,1\}^n$ is identified with the tape word
$w'\in\{1,3\}^n$ defined by  $w'_i = 2 w_i + 1$.

A \emph{configuration} of $M$ is a triple $(u, q, v)$ where $q \in Q$
is the current state and $u, v \in \Gamma^\omega$ encode the left and
right halves of the (bi-infinite) tape, with the head positioned at
the leftmost symbol of $v$. The set of configurations is denoted
$\CONFIGURATIONSMT$. The transition function $\delta$ induces the
single-step map $\step_M \colon \CONFIGURATIONSMT \to 
\CONFIGURATIONSMT$. A run on input $w$ is the sequence of 
configurations $c_0, c_1, c_2, \dots$ where $c_0 = (B^\omega, q_0, 
w \cdot B^\omega)$ and $c_{t+1} = \step_M(c_t)$. We say $M$ 
\emph{accepts $w$ in $t$ steps} if $c_t$ has state $q_F$.

\begin{remark}
We will sometimes consider $M$ acting on an \emph{infinite binary 
tape} $\tau \in \{0,1\}^\omega$ instead of a finite input; the initial
configuration is then $c_0(\tau) = (B^\omega, q_0, \tau' \cdot 
B^\omega) \to \tau'$ where $\tau' = (2\tau_1 + 1)(2\tau_2 + 1)\cdots 
\in \{1,3\}^\omega$. Within $T$ steps $M$ visits at most $T$ tape
positions, so a finite input $w$ of length $\ge T$ behaves identically
to any infinite extension $w \cdot \tau'$.
\end{remark}

\subsection{Encoding configurations}

Let $\onethreefouradic$ denote the set of base-$4$ reals with digits in
$\{1,3\}$, that is,
\[
   \onethreefouradic
   \;=\;
   \Bigl\{ \sum_{i=1}^{\infty} a_i\,4^{-i} \;:\; a_i \in \{1,3\} \Bigr\}
   \;\subseteq\; [0,1].
\]
This is a Cantor-like fractal subset of $[0,1]$. The choice of digits in
$\{1,3\}$ rather than $\{0,1,2,3\}$ creates uniform gaps between the
leading-digit cylinders, which is what allows continuous gadgets to
discriminate them. This trick is inspired from \cite{SS91b}: see also discussions and references in Remark \ref{rk18}.

A configuration $C \in \CONFIGURATIONSMT$ of a Turing machine $M$
over alphabet $\Gamma = \{1,3,B\}$ (with blank $B$) is a triple
$(u, q, v)$ where $q$ is a state and $u, v \in \Gamma^\omega$ are
the left and right halves of the tape. Following~\cite{MFCS2025},
we encode such a configuration into \emph{two} reals by a map
\[
  \nu \colon \CONFIGURATIONSMT \;\longrightarrow\;
  ([0,1] \cap \overline{\onethreefouradic})^2,
  \qquad
  \nu(C) = \bigl(\enc_{\{1,3\}}(u),\;\enc_M(q, v)\bigr),
\]
where the second coordinate \emph{co-encodes} state and right tape
in base $4$: $\enc_M(q, v)$ places a constant-length state prefix
$\enc_M(q)$ of length $|q|_4 := \lceil \log_4 |Q| \rceil$ ahead of
$\enc_{\{1,3\}}(v)$. This packing keeps the working dimension at
$2$; the equivalent ``naive'' encoding $(u, v, q)$ would require
$3$ dimensions. The width-$4$ network of \Cref{thm:main} embeds
$\nu(C)$ into its first two coordinates, the third being a scratch
register and the fourth a validity register
(\Cref{sec:def}).

\subsection{Auxiliary tool: efficient compilation of one-dimensional CPAMs}

We need to write some particular continuous affine maps (CPAM) as particular narrow neural networks.  

\subsubsection{Some known statements}

We review some known statements.  We improve some of them, as we need to discuss very small width constructions, and also some effectivity considerations.

\begin{theorem}[\cite{arora2016understanding}] \label{th-nn-pam} Every CPAM $f: \mathbb{R}^d \rightarrow \mathbb{R}$ can be represented by a  $\Netnotatione{\ReLU}{d}{1}$.
\end{theorem}
Note that this is easily extended to CPAMs with any finite output dimension.
However, in practice, the question of the width or depth of the involved  $\ReLU$-net that performs a CPAM function is relevant.   The article~\cite{hanin2017approximating} studies the numbers $w_{\min }\left(d_{i n}, d_{o u t}\right)$, defined to be the minimal value of  width $w$ such that for every continuous function $f:[0,1]^{d_{i n}} \rightarrow \mathbb{R}^{d_{o u t}}$ and every $\varepsilon>0$ there is a $\operatorname{ReLU}$ net $\NN$ with input dimension $d_{i n}$, hidden layer widths at most $w$, and output dimension $d_{\text {out }}$ that $\varepsilon$-approximates $f$. %
The main result of~\cite{hanin2017approximating} is the following estimate for $w_{\min }\left(d_{\text {in }}, d_{\text {out }}\right)$.

\begin{theorem}[{\cite[Theorem 1]{hanin2017approximating}}]  \label{th:un}
For every $d_{\text {in }}, d_{\text {out }} \geq 1$,
$
d_{\text {in }}+1 \leq w_{\min }\left(d_{\text {in }}, d_{\text {out }}\right) \leq d_{\text {in }}+d_{\text {out }} .
$
\end{theorem}

The following has been discussed:

\begin{definition}[Max-min string, {\cite[Definition 1]{hanin2017approximating} }] \label{def7}
A function $g: \mathbb{R}^{d_{i n}} \rightarrow \mathbb{R}^{d_{o u t}}$ is a max-min string of length $L \geq 1$ on $d_{\text {in }}$ input variables and $d_{\text {out }}$ output variables if there exist affine functions $\lin_1, \ldots, \lin_L$ : $\mathbb{R}^{d_{\text {in }}} \rightarrow \mathbb{R}^{d_{\text {out }}}$ such that
$$
g=\sigma_{L-1}\left(\lin_L, \sigma_{L-2}\left(\lin_{L-1}, \ldots, \sigma_2\left(\lin_3, \sigma_1\left(\lin_1, \lin_2\right)\right) \cdots\right)\right.
$$
where each $\sigma_i$ is either a coordinate-wise max or a min.
\end{definition}

\begin{proposition}[Width efficient computation of max-min string, {\cite[Proposition 2]{hanin2017approximating}}]\label{prop2haninbis} 
For every max-min string $g$ on $d_{\text {in}}$ input variables and $d_{\text {out}}$ output variables with length $L$ and every compact $K \subseteq \mathbb{R}^{d_{i n}}$, there exists a $\Netnotation{\ReLU}{d_{in}}{d_{out}}{d_{in}+d_{out}}$   and depth $L$ that computes $x \mapsto g(x)$ for every $x \in K$.
\end{proposition}

\subsubsection{What we  actually use}

For our purposes, we only need the well-known fact that any 
one-dimensional continuous piecewise-affine map (CPAM) is expressible
as a ReLU expression via the standard hinge representation. The 
specific gadgets used in the present paper (\Cref{lem:err,lem:lead,lem:shift})
are constructed explicitly with width-$2$ implementations that do not
appeal to a general width-optimal compilation. They can be seen as (sometimes optimized) particular instances of the following observation.

\begin{lemma}[Compilation of bounded CPAMs]
\label{lem:cpam}
Let $M > 0$ and let $f \colon [0, M] \to \RP$ be a continuous piecewise
affine map with $n$ breakpoints $0 < a_1 < a_2 < \cdots < a_n \le M$.
Then $f$ is computed exactly on $[0, M]$ by a $\ReLU$ network of
width $2$ and depth $2n + 2$ (including one affine output layer).
\end{lemma}

\begin{proof}
Let $\alpha_0, \ldots, \alpha_n$ denote the slopes of $f$ on the
subintervals $[0, a_1], [a_1, a_2], \ldots, [a_n, M]$, set $\beta_0 := f(0)$,
and adopt the convention $a_0 := 0$. The hinge decomposition (see e.g.~\cite{arora2016understanding}) reads
\[
  f(x) \;=\; \beta_0 + \alpha_0 x \;+\; \sum_{i=1}^{n} (\alpha_i - \alpha_{i-1})\, \ReLU(x - a_i),
  \qquad x \in [0, M].
\]
Define the partial sums $S_j(x) := \beta_0 + \alpha_0 x + \sum_{i=1}^{j} (\alpha_i - \alpha_{i-1}) \ReLU(x - a_i)$
for $j \in \{0, \ldots, n\}$, so that $S_0(x) = \beta_0 + \alpha_0 x$ and $S_n = f$.

\paragraph{Positivity offset.}
Each $S_j$ is continuous on the compact $[0,M]$, hence
\[
  C \;:=\; \max_{0 \le j \le n}\ \max_{x \in [0, M]}\ \max\!\bigl(0,\, -S_j(x)\bigr)
\]
is finite (with explicit bound $C \le |\beta_0| + L M$, where
$L := |\alpha_0| + \sum_{i=1}^{n} |\alpha_i - \alpha_{i-1}|$).
Set $\widetilde S_j(x) := S_j(x) + C \ge 0$ and $u_j(x) := \ReLU(x - a_j)$,
so $u_0(x) = x$.

\paragraph{Construction.}
We build a width-$2$ network whose state after the $j$-th \emph{block}
of two ReLU layers equals $(u_j(x),\, \widetilde S_j(x))$.

\smallskip
\emph{Initialization (1 layer).} The input $x \in [0, M]$ is lifted by
\[
  x \;\longmapsto\; \ReLU\!\begin{pmatrix} x \\ \alpha_0 x + \beta_0 + C \end{pmatrix}
  \;=\; \begin{pmatrix} u_0(x) \\ \widetilde S_0(x) \end{pmatrix},
\]
the equality holding because both coordinates are $\ge 0$ on $[0,M]$
(the first by hypothesis, the second by choice of $C$).

\smallskip
\emph{Block $j \to j+1$ (2 layers).}
\textsc{Sub-layer A.} Apply
\[
  \begin{pmatrix} u_j \\ \widetilde S_j \end{pmatrix}
  \;\longmapsto\;
  \ReLU\!\begin{pmatrix} u_j - (a_{j+1} - a_j) \\ \widetilde S_j \end{pmatrix}.
\]
The second coordinate is preserved since $\widetilde S_j \ge 0$. The first
equals $u_{j+1}(x)$: on $\{x \ge a_j\}$, $u_j(x) = x - a_j$ so
$\ReLU(u_j - (a_{j+1}-a_j)) = \ReLU(x - a_{j+1}) = u_{j+1}(x)$; on
$\{x < a_j\}$, $u_j(x) = 0$ and $\ReLU(-(a_{j+1}-a_j)) = 0 = u_{j+1}(x)$
(since $x < a_j < a_{j+1}$).

\textsc{Sub-layer B.} Apply
\[
  \begin{pmatrix} u_{j+1} \\ \widetilde S_j \end{pmatrix}
  \;\longmapsto\;
  \ReLU\!\begin{pmatrix} u_{j+1} \\ \widetilde S_j + (\alpha_{j+1} - \alpha_j)\, u_{j+1} \end{pmatrix}
  \;=\; \begin{pmatrix} u_{j+1} \\ \widetilde S_{j+1} \end{pmatrix}.
\]
The first coordinate is preserved since $u_{j+1} \ge 0$. The second
argument equals $\widetilde S_{j+1}(x)$, which is $\ge 0$ by definition
of $C$; hence the ReLU acts as the identity.

\smallskip
\emph{Output (1 affine layer).} After $2n$ ReLU layers in blocks, the
state is $(u_n(x),\, \widetilde S_n(x)) = (u_n(x),\, f(x) + C)$. The affine
map $(u, v) \mapsto v - C$ yields $f(x)$.

\paragraph{Width and depth.}
Width is $2$ throughout; depth is $1 + 2n + 1 = 2n + 2$.
\end{proof}

\subsection{Width-3 step simulation}

\begin{theorem}[Simulation of TM, {\cite[Theorem~23]{MFCS2025}}]
\label{thm:simulation-optimize}
Let $M$ be a Turing machine with input alphabet $\Sigma=\{1,3\}$ and tape
alphabet $\Gamma=\{1,3,B\}$, and let $\CONFIGURATIONSMT
=\Gamma^\omega \times Q \times \Gamma^\omega$ be its configuration space.
There exist a piecewise-affine function
$g_M \colon \R^2 \to \R^2 \in \Netnotation{\ReLU}{2}{2}{3}$ and an
encoding map $\nu \colon \CONFIGURATIONSMT \to
([0,1] \cap \overline{\onethreefouradic})^2$ such that the diagram below
commutes:
\[
\begin{CD}
\CONFIGURATIONSMT @>{\vdash}>> \CONFIGURATIONSMT \\
@V{\nu}VV  @VV{\nu}V \\
\R^2 @>{g_M}>> \R^2
\end{CD}
\]
That is, $g_M(\nu(C)) = \nu(C')$ for every pair of configurations
$C \vdash C'$ in $\CONFIGURATIONSMT$.
\end{theorem}

\begin{proof}
Since the original argument in~\cite{MFCS2025} leaves several
technical points implicit, we give a self-contained derivation,
combining a structural decomposition of the TM step with the
one-dimensional compilation of Lemma~\ref{lem:cpam}.

\paragraph{Decomposition into one-coordinate atomic updates.}
The transition $\vdash$ of \cite{MFCS2025} acts on the encoded pair $\nu(C) \in [0,1]^2$
through a finite sequence of atomic operations: read the leading
base-$4$ digit of the active half-tape (Lemma~\ref{lem:lead}), branch on
the state-symbol pair via a constant-depth case split on the finitely
many such pairs, write the new symbol, shift the head, update the
state prefix. Each atomic operation modifies \emph{exactly one} of
the two encoded coordinates $y_1 = \enc_{\{1,3\}}(u)$,
$y_2 = \enc_M(q, v)$ and reduces, after the case split, to a
continuous piecewise-affine map of one variable with $O_M(1)$
breakpoints (the gadgets of~Lemmas~\ref{lem:lead} and \ref{lem:shift} are particular
instances). Since the number of atomic operations depends only on
$M$, the construction yields
$$
   g_M = \phi_K \circ \cdots \circ \phi_1,
$$
where $K$ depends on $M$ and each $\phi_i \colon \R^2 \to \R^2$ has the form
$(y_1, y_2) \mapsto (y_1,\, \psi_i(y_2))$, or $(y_1, y_2) \mapsto (\psi_i(y_1),\, y_2)$, with $\psi_i$ a 1D CPAM
taking values in $[0,1]$.

\paragraph{One atomic update fits in width~$3$.}
Take the first form (the second is symmetric). Lemma~\ref{lem:cpam}
provides a width-$2$ $\ReLU$ subnetwork computing $\psi_i$,
internally maintaining a $2$-register state (staircase variable and
shifted partial sum). We place these two registers in
coordinates~$2$ and~$3$ of the ambient width-$3$ network, while
coordinate~$1$ carries $y_1$ through every $\ReLU$ layer via
$\ReLU(y_1) = y_1$ (valid since $y_1 \in [0,1] \subseteq \R_{\ge 0}$).
A constant-depth $\ReLU$ junction at the end of the block extracts
$\psi_i(y_2)$ into
coordinate~$2$, zeros coordinate~$3$, and preserves $y_1$ in
coordinate~$1$, restoring the layout $y_1,\, \psi_i(y_2),\, 0)$
required at the entry of $\phi_{i+1}$. (See Figure~\ref{fig:psi-block} for an illustration).
\input{FIG/fig-TM}

\paragraph{Concatenation.}
Stacking the $K$ blocks produces $g_M \in \Netnotation{\ReLU}{2}{2}{3}$
of depth $\sum_{i=1}^{K} O(n_i) = \mathrm{poly}(|M|)$, where $n_i$ is
the number of breakpoints of $\psi_i$. The commutation
$g_M \circ \nu = \nu \circ {\vdash}$ on $\CONFIGURATIONSMT$ holds
cylinder-by-cylinder by the explicit case analysis
in~\cite{MFCS2025}; outside the encoded domain
$\nu(\CONFIGURATIONSMT)$, $g_M$ is the continuous piecewise-affine
extension determined by the construction, with no claim made.
\end{proof}

\begin{remark}[Equivalence with \Cref{thm:width3-step}]
\label{rem:dim2-to-width3}
\Cref{thm:width3-step} in the main body differs from
\Cref{thm:simulation-optimize} only in its bookkeeping: the scratch
register is exposed as an explicit third input/output coordinate,
reset to $0$ between TM steps. The conversion from one formulation to the other is purely affine and does
not increase the hidden width: prepend an $\R^3 \to \R^3$ map dropping the
third input coordinate, and append an $\R^3 \to \R^3$ map padding the
output with a zero third coordinate. We use the two-dimensional formulation
for the width analysis above and the three-dimensional register
form when chaining the simulator with the encoding-validity
certificate of Stage~2 in \Cref{sec:main-proof-sketch}.
\end{remark}

\begin{remark}\label{rk18}
The use of base-$4$ digits in $\{1,3\}$ rather than a binary
representation is essential: a continuous $\ReLU$ map cannot
discriminate adjacent base-$2$ digits, but it can discriminate
$\{1,3\}$-digits because of the gap of size $\tfrac{1}{2}$ between
$\frac{1}{4}$ and $\frac{3}{4}$. See~\cite{BBKPC00,Moo91,KCG94,BC96,SS91b}
for related work.
Note that the gaps are between the leading-digit cylinders, not
between admissible points: $\onethreefouradic$ is a perfect set, two
encodings agreeing on their first $m$ digits differ by at most
$4^{-m}$, so no fixed margin separates distinct encodings and no
rounding step back onto the encoding set exists. This is one more
reason why uniform approximation does not yield exact simulation
(the main one is the expansion argument of
\Cref{sec:main-sketch-obstruction}).
\end{remark}

\section{Encoding-validity gadget at width 4}
\label{app:encoding}

\Cref{thm:simulation-optimize} requires a validly encoded input from
$\onethreefouradic$. To make the simulation work uniformly on any input in
$[0,1]^4$, we construct a width-$4$ preprocessing gadget that detects
deviation from the encoding set. This appendix develops the auxiliary
gadgets and combines them.

\subsection{Why we need this gadget}
\label{app:encoding-why}

If width and dimension were unrestricted, $\NP$-completeness of
verification for $\ReLU$ networks would follow from a SAT-style reduction
\citep{wurm1}, or from the bounded-time reachability hardness for
two-dimensional CPAMs~\citep{RP14}. Translating the CPAM to a $\ReLU$
network via~\citep{arora2016understanding} yields verification hardness for
$\ReLU$ networks of some fixed width, but the width is not estimated and is
at best loose. Moreover, this approach is unrelated to Turing-machine
complexity, so it gives only $\NP$-hardness rather than the full
time-bound spectrum (\Cref{thm:verif-spectrum}).

We instead exploit \Cref{thm:simulation-optimize}: $\ReLU$ networks of
width $3$ already simulate Turing machines, but on the fractal encoding
$\onethreefouradic$. Since the verification problem asks for the existence
of an input in a \emph{convex} region $\Init$, we must close the gap between
``valid encoding inputs'' and ``arbitrary convex inputs''. The
encoding-validity gadget below performs this closure with one extra unit
of width.

\subsection{Auxiliary gadgets}
\label{app:encoding-gadgets}

We construct three width-$2$ subnetworks that constitute the building
blocks of the validity check.

\begin{lemma}[Error indicator]
\label{lem:err}
There exists a $\ReLU$ network $\NNErr \colon \R \to \R^2$ in
$\Netnotation{\ReLU}{1}{2}{2}$ such that, for every $x \in [0,4]$,
\(
   \NNErr(x) = (x, \Err(x)),
\)
where $\Err \colon \R \to \R_{\ge 0}$ satisfies $0 \le \Err(x) \le 1$ for
all $x \in \RP$, and
\(
   \Err(x) = 0 \iff x \in [1,2] \cup [3,4].
\)
\end{lemma}

\begin{proof}
Let $f_{(0,1)}(x) = \ReLU(1-x)$ and
$f_{(2,3)}(x) = \tfrac{1}{2} - \ReLU(x - \tfrac{5}{2})
- \ReLU(\tfrac{5}{2} - x)$. Then $f_{(0,1)}$ is strictly positive on $[0,1)$
and non-positive elsewhere; $f_{(2,3)}$ is strictly positive on $(2,3)$ and
non-positive elsewhere. Define
\(
   \Err(x) = \ReLU(f_{(0,1)}(x)) + \ReLU(f_{(2,3)}(x)).
\)
By construction, $\Err(x) = 0$ iff $x \in [1,2] \cup [3,4]$. Rewriting as
a nested $\ReLU$ composition gives
\[
   \Err(x)
   = \ReLU\!\Bigl(
       3 - x - \ReLU\!\bigl(
                  5 - 2x - \ReLU\!\bigl(
                              2 - x + \ReLU(1-x)
                            \bigr)
                \bigr)
     \Bigr),
\]
which is a finite composition of $\ReLU$ and affine operations on 
a single real input, hence directly implementable by a $\ReLU$ 
network of width $2$: at each layer one neuron computes the inner 
$\ReLU$ expression while the second neuron carries the running 
input value $x$ via $\ReLU(x) = x$ (valid since $x \ge 0$ on the 
domain $[0,4]$). See Figure~\ref{fig:err}.

\input{./FIG/fig-err}
\end{proof}

\begin{lemma}[Leading-digit extraction]
\label{lem:lead}
There exists a $\ReLU$ network $\NNlead \colon \R \to \R^2$ in
$\Netnotation{\ReLU}{1}{2}{2}$ such that, for every
$x = \sum_{i=1}^{\infty} a_i 4^{-i} \in (0,1)$,
\(
   \NNlead(x) = (x, \lead(x)),
\)
where $\lead(x) = a_1$ whenever $a_1 \in \{1,3\}$.
\end{lemma}

\begin{proof}
The digits $a_1 \in \{1,3\}$ correspond exactly to
$x \in [\frac{1}{4}, \frac{1}{2}] \cup [\frac{3}{4}, 1]$ (with gaps in
between). Define
\[
   \lead(x) = 1 + 8 \ReLU(x - \tfrac{1}{2}) - 8 \ReLU(x - \tfrac{3}{4}).
\]

\input{FIG/fig-lead}

A direct case analysis  verifies:
$x \in [\tfrac{1}{4}, \tfrac{1}{2}] \Rightarrow \lead(x) = 1$;
$x \in [\tfrac{3}{4}, 1] \Rightarrow \lead(x) = 3$. See Figure~\ref{fig:lead}.
\end{proof}

\begin{lemma}[Base-$4$ digit shift]
\label{lem:shift}
There exists a $\ReLU$ network $\NNshift \colon \R \to \R^2$ in
$\Netnotation{\ReLU}{1}{2}{2}$ such that, for every $x \in (0,1)$,
\(
   \NNshift(x) = (x, \shift(x)),
\)
where $\shift(x) \in [0,1]$ for $x \in (0,1)$ and
\(
   \shift(x) = \sum_{i=2}^{\infty} a_i\,4^{-(i-1)}
\)
when $x = \sum_{i=1}^{\infty} a_i 4^{-i}$ with $a_1 \in \{1,3\}$.
\end{lemma}

\begin{proof}
It suffices to verify $\shift(x) = 4x - \lead(x)$. %
\end{proof}

\subsection{Width-$4$ validity check via the $\{1,3\}$-base-$4$ digit gadget}
\label{app:encoding-check}

\begin{proposition}[Width-$4$ encoding-validity check]
\label{prop:checking-4}
Fix $n \in \N$. There exists a $\ReLU$ network
$\NNT{n} \in \Netnotation{\ReLU}{2}{2}{4}$ such that, for every
$x = \sum_{i=1}^{\infty} a_i 4^{-i} \in [0,1]$,
\(
   \NNT{n}(x, 0) = (T_n(x), \varepsilon_n(x))
\)
satisfies:
\begin{enumerate}
  \item if $x = \sum_{i=1}^{n} a_i 4^{-i}$ with $a_i \in \{1,3\}$,
        then $T_n(x) = x$ and $\varepsilon_n(x) = 0$;
  \item conversely, if $\varepsilon_n(x) = 0$, then $x$ has the form
        $\sum_{i=1}^{n} a_i 4^{-i}$ with $a_i \in \{1,3\}$.
\end{enumerate}
\end{proposition}

\begin{proof}
The network $\NNT{n}$ iteratively decodes the first $n$ base-$4$ digits of
$x$, accumulating any deviation from $\{1,3\}$ into an error variable
$\epsilon$.

\medskip
\noindent\textbf{Construction.} Given the input pair $(x, p)$, the network
performs the following loop:
\begin{enumerate}
  \item Initialise $\epsilon \gets 0$.
  \item For $i = 1, \dots, n$:
    \begin{enumerate}
      \item Update the error: $\epsilon \gets \ReLU(\epsilon + \Err(4x))$.
      \item Extract the leading base-$4$ digit: $a \gets \lead(x)$.
      \item Update the accumulator: $p \gets \ReLU(p + a \cdot 4^{-i})$.
      \item Shift the input: $x \gets \shift(x)$.
    \end{enumerate}
  \item Return $(p, x + \epsilon)$.
\end{enumerate}

By Lemmas~\ref{lem:err}, \ref{lem:lead}, and~\ref{lem:shift}, each
instruction can be implemented by a width-$2$ $\ReLU$ subnetwork. The
registers $p$ and $\epsilon$ must be propagated throughout the computation
alongside the working register $x$, yielding total width $4$; see
Figure~\ref{fig:SKIP-Trunck-4}.

\medskip
\noindent\textbf{Correctness.} Let $(x_i, p_i, \epsilon_i)$ denote the
values after $i$ iterations, with initial values $(x_0, p_0, \epsilon_0)
= (x, 0, 0)$.

\emph{Property 1.} Suppose $x = \sum_{j=1}^{n} a_j 4^{-j}$ with
$a_j \in \{1,3\}$. By induction, $x_i = \sum_{j=i+1}^{\infty} a_j 4^{-(j-i)}$
and $p_i = \sum_{j=1}^{i} a_j 4^{-j}$. After $n$ iterations, $x_n = 0$ and
$p_n = x$. Moreover, $\epsilon_n = \sum_{i=1}^{n} \Err(4 x_{i-1}) = 0$
since each $\Err(4 x_{i-1})$ vanishes when $a_i \in \{1,3\}$ by
Lemma~\ref{lem:err}. Hence $T_n(x) = p_n = x$ and $\varepsilon_n(x) = x_n + \epsilon_n = 0$.

\emph{Property 2.} Suppose $\NNT{n}(x, 0) = (x', 0)$, i.e.
$x_n + \epsilon_n = 0$. Both $x_n \ge 0$ (since $\shift$ is non-negative)
and $\epsilon_n \ge 0$ (sum of $\ReLU$s), so $x_n = 0$ and $\epsilon_n = 0$.
We argue by contradiction: suppose $x$ is not of the form
$\sum_{j=1}^{n} a_j 4^{-j}$ with $a_j \in \{1,3\}$. Two cases.
\begin{itemize}
  \item The first $n$ digits are in $\{1,3\}$, but
        $x \neq \sum_{j=1}^{n} a_j 4^{-j}$. Then
        $x_n = \sum_{j > n} a_j 4^{-(j-n)} > 0$, contradiction
        (on $\onethreefouradic$, $x_n \ge 1/3$).
  \item There is some $\ell \le n$ with $a_\ell \notin \{1,3\}$, taken
        minimal. Then $\Err(4 x_{\ell-1}) > 0$, hence $\epsilon_n > 0$,
        contradiction.
\end{itemize}
This completes the proof.
\input{./FIG/fig-Trunck-4}
\end{proof}

\begin{remark}[The per-step certificate $\Cert$ of \Cref{thm:main}]
\label{rem:checking-applies-twice}
The per-step certificate of \Cref{thm:main} is
\[
   \Cert(x_1, x_2) \;=\; \max\bigl\{\, \Err(4 x_1),\;
       \Err\bigl(4\,\mathrm{suf}(x_2)\bigr),\;
       d_Q\bigl(\mathrm{pre}(x_2)\bigr) \,\bigr\} \;\in\; [0,1],
\]
where $\mathrm{pre}(x_2)$ and $\mathrm{suf}(x_2)$ are the
state-prefix and tape-suffix portions of $x_2$ (\Cref{sec:enc}),
obtained by a constant-depth slicing (a CPAM, expressible as a
finite ReLU expression by Lemma~\ref{lem:cpam}), and
$d_Q \colon [0,1] \to [0,1]$ is a state-distance subgadget that
vanishes exactly on $\enc_M(Q)$, built as a $|Q|$-fold minimum of
tent functions. The maxima are realized in width $2$ via
$\max\{a, b\} = a + \ReLU(b - a)$. Thus $\Cert(x_1, x_2) = 0$ iff
the leading base-$4$ digit of $x_1$ and of the tape-suffix of $x_2$
lies in $\{1,3\}$ (that is, $4 x_1$ and $4\,\mathrm{suf}(x_2)$ lie
in the closed cylinders $[1,2] \cup [3,4]$, Lemma~\ref{lem:err})
and the state prefix lies in $\enc_M(Q)$. The zero set of $\Cert$ is
a finite union of boxes, which a finite $\ReLU$ network can realize.
Membership of the \emph{full} expansions in $\onethreefouradic$
cannot be checked in one step (\Cref{prop:obstruction}): it is
checked across iterations, one digit per step, because the
simulator's own tape shifts expose successive digits as leading
digits. The gadget of Proposition~\ref{prop:checking-4} belongs to the
finite-word setting of~\cite{FSTTCS2026}: it vanishes exactly when
$x$ is a base-$4$ word of length $n$ over $\{1,3\}$, and is positive
on $\onethreefouradic$. Only its one-digit building block
$\Err$ (Lemma~\ref{lem:err}) enters $\Cert$.
\end{remark}

\section{Full proof of \texorpdfstring{\Cref{thm:main}}{the main theorem}}
\label{app:main-proof}

This appendix formalizes the construction sketched in
\Cref{sec:main-proof-sketch} of the main body,  using the building blocks of
\Cref{app:width3,app:encoding}.  We construct $N_M$
explicitly, establish a structural invariant on the certificate,
and derive parts~(i) and~(ii) of \Cref{thm:main}.

\begin{remark}[Certifying a set]
\label{rem:certify-app}
Part~(ii) of \Cref{thm:main} says, informally, that $N_M$
\emph{certifies} a set $V \subseteq [0,1]^4$: a coordinate is
non-decreasing along the iterates and vanishes along the whole
trajectory of $x$ exactly when $x \in V$. Here $V$ is the set of
inputs whose visited digits (\Cref{sec:enc}) lie in $\{1,3\}$ and
whose state prefix stays in $\enc_M(Q)$. This set contains the
encoded configurations, and it is exactly what soundness requires.
The quantification over all $k$ is essential, since no finite $\ReLU$
network has zero set $\onethreefouradic$ (\Cref{rem:online-cert}).
The certificate appears only inside the witness network built for
hardness (\Cref{rem:certify}).
\end{remark}

\subsection{Construction of the simulating network $N_M$ for a Turing machine $M$}
\label{app:nm-construction}

Recall the encoding map from \Cref{app:width3}: a configuration 
$c = (u, q, v) \in \CONFIGURATIONSMT$ is encoded as 
$\nu(c) = (\enc_{\{1,3\}}(u), \enc_M(q, v)) \in [0,1]^2$, where the
second coordinate co-encodes state and right tape. The full input 
to $N_M$ is
\[
   \enc_M(c) := \bigl(\nu(c)_1,\; \nu(c)_2,\; 0,\; 0\bigr) 
   \;\in\; [0,1]^4,
\]
with the third coordinate used as scratch by the width-$3$ simulator
$g_M$ and the fourth as a validity accumulator.

The network $N_M \colon [0,1]^4 \to [0,1]^4$ is defined as the 
composition $N_M = N_{\mathrm{sim}} \circ N_{\mathrm{check}}$ of 
two register-network blocks; see Figures~\ref{fig:width4-schematic}
(schematic) and~\ref{fig:register} (formal).
\begin{enumerate}
\item[(C)] $N_{\mathrm{check}}$ \emph{(width-$4$ certificate update)}:
\[
   (x_1, x_2, x_3, x_4) \;\longmapsto\;
   \bigl(x_1,\; x_2,\; x_3,\; \max\{x_4,\, \Cert(x_1, x_2)\}\bigr),
\]
where $\Cert \colon [0,1]^2 \to [0, 1]$ is the \emph{per-step}
certificate of Remark~\ref{rem:checking-applies-twice}:
$\Cert(x_1, x_2) = 0$ iff the leading base-$4$ digit of $x_1$ and of
the tape-suffix of $x_2$ lies in $\{1,3\}$ and the state prefix of
$x_2$ lies in $\enc_M(Q)$. Membership of the full expansions in
$\onethreefouradic$ is checked across iterations, one digit per step
(Remark~\ref{rem:checking-applies-twice}). The $\max$ is realized in
width $2$ via $\max\{a, b\} = a + \ReLU(b - a)$.

\item[(S)] $N_{\mathrm{sim}}$ \emph{(width-$3$ simulation)}: applies
the width-$3$ network $g_M \in \Netnotation{\ReLU}{2}{2}{3}$ of
\Cref{thm:simulation-optimize} to $(x_1, x_2)$ using $x_3$ as 
scratch, while propagating $x_4$ unchanged through an identity wire
(this is consistent with the register-form structure of $N_M$, see
Remark~\ref{rem:nm-register}).
\end{enumerate}
\input{FIG/fig-register}
The order (check \emph{before} simulation) is essential: any
encoding violation is detected before $g_M$ has the chance to
overwrite $(x_1, x_2)$. The total per-iteration depth is 
$D_M = O(\mathrm{depth}(g_M)) = \mathrm{poly}(|M|)$, and the width is~$4$.

\subsection{Structural invariant}
\label{app:invariant}

The key technical content of the proof is the following invariant,
which formalizes the claim made informally in 
Stage~2 of \Cref{sec:main-proof-sketch}.

\begin{lemma}[Encoding-validity invariant]
\label{lem:invariant}
Let $x = (x_1^{(0)}, x_2^{(0)}, x_3^{(0)}, x_4^{(0)}) \in [0,1]^4$
be any input, and let
$(x_1^{(k)}, x_2^{(k)}, x_3^{(k)}, x_4^{(k)}) := N_M^{[k]}(x)$ for
$k \ge 0$. Call step $j$ \emph{locally valid} if the leading
base-$4$ digit of $x_1^{(j)}$ and of the tape-suffix of $x_2^{(j)}$
lies in $\{1,3\}$ and the state-prefix of $x_2^{(j)}$ lies in
$\enc_M(Q)$. Then for every $k \ge 0$:
\begin{enumerate}
\item[(a)] (Monotonicity.) $x_4^{(k+1)} \ge x_4^{(k)}$ and
$x_4^{(k)} \in [0, 1]$.

\item[(b)] (Local validity preserves the certificate.) If every step
$j \le k$ is locally valid, then $x_4^{(k)} = x_4^{(0)}$. In
particular, if $x_1^{(0)} \in \onethreefouradic$, the tape-suffix of
$x_2^{(0)}$ is in $\onethreefouradic$ and the state-prefix of
$x_2^{(0)}$ is in $\enc_M(Q)$, then every step is locally valid,
since $g_M$ preserves these properties
(\Cref{thm:simulation-optimize}), and $x_4^{(k)} = x_4^{(0)}$ for
all $k$.

\item[(c)] (Violation detected.) Conversely, if some step $j \le k$
is not locally valid, then
$x_4^{(k)} \ge \Cert(x_1^{(j)}, x_2^{(j)}) > 0$ for all
$k \ge j + 1$. In particular, if $x_4^{(0)} = 0$, then
$x_4^{(k)} > x_4^{(0)}$, and this persists forever.
\end{enumerate}
Summed over the steps $j < k$, the local-validity conditions are
exactly the condition of \Cref{thm:main}(ii): every digit visited
during the first $k$ iterations (\Cref{sec:enc}) lies in $\{1,3\}$,
and the state prefix stays in $\enc_M(Q)$ along the trajectory.
\end{lemma}

\begin{proof}
(a) The fourth coordinate is updated by $x_4^{(k+1)} =
\max\{x_4^{(k)}, \Cert(x_1^{(k)}, x_2^{(k)})\}$, which is
non-decreasing. Since $\Cert \le 1$
(Remark~\ref{rem:checking-applies-twice}) and
$x_4^{(0)} \in [0,1]$, induction gives $x_4^{(k)} \in [0,1]$.

(b) If step $j$ is locally valid, then
$\Cert(x_1^{(j)}, x_2^{(j)}) = 0$ by
Remark~\ref{rem:checking-applies-twice}, hence
$x_4^{(j+1)} = \max\{x_4^{(j)}, 0\} = x_4^{(j)}$. Iterating from
$j = 0$ to $k - 1$ yields $x_4^{(k)} = x_4^{(0)}$. The second claim
follows since $g_M$ maps valid encodings to valid encodings
(\Cref{thm:simulation-optimize}), so every leading digit exposed
along the trajectory stays in $\{1,3\}$ and the state prefix stays
in $\enc_M(Q)$.

(c) If step $j$ is not locally valid, then
$\Cert(x_1^{(j)}, x_2^{(j)}) > 0$
(Remark~\ref{rem:checking-applies-twice}), so
$x_4^{(j+1)} \ge \Cert(x_1^{(j)}, x_2^{(j)}) > 0$. By monotonicity,
$x_4^{(k)} \ge x_4^{(j+1)}$ for all $k \ge j + 1$.
\end{proof}

\subsection{Proof of \Cref{thm:main}}

\begin{proof}[Proof of \Cref{thm:main}]
\emph{Part (i): faithful simulation on encoded inputs.}
Let $\tau \in \{0,1\}^\omega$ be an infinite binary tape and 
$c_0 := c_0(\tau)$ the initial configuration of $M$ on $\tau$. The
input is $\enc_M(c_0) = (\nu(c_0)_1, \nu(c_0)_2, 0, 0)$ with both
$\nu(c_0)_1$ and $\nu(c_0)_2$ encoded as required.

We claim by induction on $t \ge 0$ that 
$N_M^{[C_M\, t]}(\enc_M(c_0)) = \enc_M(c_t(\tau))$, where $C_M$ is
the per-step iteration count of $g_M$ in 
\Cref{thm:simulation-optimize}. The base case $t = 0$ is immediate.
For the inductive step, suppose $N_M^{[C_M\, t]}(\enc_M(c_0)) = 
\enc_M(c_t)$ for some $t$. By \Cref{thm:simulation-optimize}, $g_M$
maps $\nu(c_t) \mapsto \nu(c_{t+1})$ in $C_M$ iterations. Since the
trajectory remains within encoded configurations (the simulator 
preserves $\onethreefouradic$-encoded tapes by 
\Cref{thm:simulation-optimize}), the validity hypothesis of
Lemma~\ref{lem:invariant}(b) holds at every step, so the fourth 
coordinate stays at $0$. Hence 
$N_M^{[C_M(t+1)]}(\enc_M(c_0)) = \enc_M(c_{t+1})$.

\smallskip
\emph{Part (ii): online certificate properties.}
The boundedness and monotonicity of $\pi_4(N_M^{[k]})$ are exactly
Lemma~\ref{lem:invariant}(a). For $x_4^{(0)} = 0$, the condition for
$\pi_4(N_M^{[k]}(x)) = 0$ to hold is exactly local validity of every
step $j < k$, which by the closing claim of
Lemma~\ref{lem:invariant} is the visited-digit condition of the
statement. The contrapositive (a violation forces $x_4 > 0$
thereafter) is Lemma~\ref{lem:invariant}(c).

\smallskip
\emph{Effective construction.}
The reduction $M \mapsto N_M$ is computable in polynomial time: 
$g_M$ is constructed in $\mathrm{poly}(|M|)$ time by 
\Cref{thm:simulation-optimize}, the certificate gadget in 
$\mathrm{poly}(|M|)$ time by Remark~\ref{rem:checking-applies-twice}, and the 
register-form composition is immediate by construction
(Remark~\ref{rem:nm-register}).
Instantiating $M = U$, where $U$ is a fixed universal TM
(Lemma~\ref{lem:tm-universal}), yields a single fixed network whose
iterates simulate every $(M', w)$.
\end{proof}

\subsection{A worked example}
\label{app:worked-example}

There is no encoder anywhere in the construction: in the
reachability problem, the input to the network is not produced by
encoding anything. It is chosen by the existential quantifier
directly inside the convex region $\Init$, and $N_M$ is applied to
whichever point the quantifier picks. Almost every point of $\Init$
is not a valid encoding, and being finitely representable is
unrelated to being a valid encoding: the rationals in $[0,1]$ are
dense, and only a measure-zero subset of them encodes anything. The
following two witnesses make this concrete. Fix any $M$ and the
regions $\Init, \Acc$ of \Cref{sec:zoo-proof}: the state prefix of
$x_2$ is pinned to $\enc_M(q_0)$, $x_3 = x_4 = 0$, and $x_1$ is free
in $[0,1]$.

\emph{Witness A (a valid encoding).} Take
$x_1 = 0.\overline{1}_4 = \tfrac13$ (all digits equal to $1$, hence
$x_1 \in \onethreefouradic$), and let the tape-suffix digits of
$x_2$ all be $1$ as well. Every step is locally valid:
$4 x_1 = \tfrac43 \in [1,2]$, so $\Err(4 x_1) = 0$
(Lemma~\ref{lem:err}), and likewise on the suffix of $x_2$; the same
happens at every later step, since $g_M$ maps encodings to encodings
(\Cref{thm:simulation-optimize}). So $x_4 = 0$ throughout and the
iterates are exactly $\enc_M(c_t)$. If $M$ accepts the all-ones
tape, the trajectory enters $\Acc$.

\emph{Witness B (not an encoding).} Take $x_1 = \tfrac18 = 0.02_4$,
with $x_2$ as in Witness~A. This point is perfectly representable
(three bits), lies in $\Init$, and is not in $\onethreefouradic$:
its leading digit is $0$. The gadget evaluates
$\Err(4 x_1) = \Err(\tfrac12)$, and with $\Err$ as in
Lemma~\ref{lem:err} the first tent contributes
$\ReLU(1 - \tfrac12) = \tfrac12$ while the second contributes $0$,
so $\Cert(x_1, x_2) = \tfrac12$. Hence $x_4$ jumps to $\tfrac12$ at
the first iteration and, being non-decreasing, stays at least
$\tfrac12$ forever. Since $\Acc$ requires $x_4 \le 0$, this witness
can never reach $\Acc$, whatever the continuous extension of $g_M$
does with it downstream.

Witness~B is exactly the case the certificate exists for: without
it, $g_M$ acting on $\tfrac18$ is an unconstrained continuous
piecewise-affine map, and nothing would prevent it from landing in
$\Acc$, making the instance answer \textsc{Yes} for reasons
unrelated to $M$.

\section{Verification spectrum at width $4$}

\subsection{Polynomial-bit-complexity witness lemma}
\label{app:np-witness}

The NP-membership claim in the proof of \Cref{thm:verif-spectrum} 
(unary $T$, width-$4$ ReLU) and in the proof of \Cref{thm:rho} 
(unary $T$, width-$6$ smooth $\rho$, after $\rho \to \ReLU$ 
emulation) rests on the following classical observation.

\begin{lemma}[Polynomial-bit-complexity for $\ReLU$ network reachability]
\label{lem:np-witness}
Let $N \colon [0,1]^d \to [0,1]^d$ be a $\ReLU$ network with rational
weights of bit complexity $b$, depth $L$, and width $w$. Let
$\Init, \Acc \subseteq [0,1]^d$ be convex regions specified by
rational linear inequalities of bit complexity $\le b'$, and let $T \in \N$.
If there exists a real $x \in \Init$ with $N^{[t]}(x) \in \Acc$ for some
$t \le T$, then there exists a rational $x \in \Init$ with the same
property and bit complexity $\mathrm{poly}(d, w, L, T, b, b')$.
\end{lemma}

\begin{proof}
Suppose $x^* \in \Init$ is a real point such that 
$N^{[t^*]}(x^*) \in \Acc$ for some $t^* \le T$. We construct a rational
witness of polynomial bit complexity.

\smallskip
\noindent\emph{Step 1: activation pattern.}
A single forward pass through a $\ReLU$ network $N$  of width $w$ and depth $L$ involves at most $wL$ neurons, hence
$wL$ $\ReLU$ activations. For each $t \in \{1, \dots, T\}$, the composition 
$N^{[t]}$ involves $wtL \le wTL$ such neurons. Hence, at most $wTL$ 
ReLU activations in total along the trajectory $x^*, N(x^*), \dots,
N^{[T]}(x^*)$. Each neuron is in one of two states: \emph{active}
($\ReLU(z) = z$, i.e. $z \ge 0$) or \emph{inactive}
($\ReLU(z) = 0$, i.e. $z \le 0$). Let 
$\sigma^* \in \{+, -\}^{wTL}$ denote the activation pattern realized
at $x^*$.

\smallskip
\noindent\emph{Step 2: polyhedral characterization.}
Fix any activation pattern $\sigma \in \{+, -\}^{wTL}$. On the set
of points $x \in [0,1]^d$ for which iterating $N$ produces the 
pattern $\sigma$, each $\ReLU$ acts as either the identity or zero,
and so $N^{[t]}$ becomes an \emph{affine} map $A_\sigma^{(t)} \colon
\R^d \to \R^d$ for every $t \le T$. The set of $x$ realizing 
$\sigma$ is the polyhedron
\[
   P_\sigma := \bigl\{ x \in [0,1]^d : 
   \mathrm{sign}(z_{i,t}(x)) = \sigma_{i,t} \text{ for every neuron } i \text{ and iteration } t \le T \bigr\},
\]
where each $z_{i,t}(x)$ is an affine function of $x$ with rational 
coefficients of bit complexity at most $\mathrm{poly}(d, w, L, T, b)$
(the bit complexity grows by at most $O(b  \log w)$ at each layer).

\smallskip
\noindent\emph{Step 3: linear system.}
For $\sigma = \sigma^*$ and $t = t^*$, the conjunction
``$x \in \Init$, $x \in P_\sigma$, $A_\sigma^{(t)}(x) \in \Acc$''
is a system $\mathcal{S}$ of rational linear inequalities. Each
inequality has rational coefficients of bit complexity 
$\mathrm{poly}(d, w, L, T, b, b')$, and the system has at most
$\mathrm{poly}(d, w, L, T, b')$ inequalities. The system $\mathcal{S}$
is feasible over $\R$ since $x^*$ satisfies it.

\smallskip
\noindent\emph{Step 4: rational witness.}
By a standard result in polyhedral combinatorics~\cite[Ch.~10]{Sch86},
any feasible system of rational linear inequalities admits a rational
solution of bit complexity polynomial in the bit complexity of the
system. Applied to $\mathcal{S}$, this yields a rational
$x \in [0,1]^d$ of bit complexity $\mathrm{poly}(d, w, L, T, b, b')$
with $x \in \Init$, $x \in P_{\sigma^*}$, and 
$N^{[t^*]}(x) = A_{\sigma^*}^{(t^*)}(x) \in \Acc$.
\end{proof}

\subsection{Underlying complexity statements about Turing machines}
\label{app:main-tm-complexity}

We use the following classical facts about Turing-machine complexity.

\begin{lemma}[Basic statements from complexity theory]
\label{lem:tm-complexity}
Let $\Sigma = \{0,1\}$. The following decision problems are complete
for the indicated complexity classes.
\begin{enumerate}
\item[(i)] (\emph{Unary time bound, $\NP$-complete.}) Given a 
deterministic Turing machine $M$ and $T \in \N$ in unary, does there
exist $w \in \Sigma^*$ and $t \le T$ such that $M$ accepts $w$ in $t$
steps?

\item[(ii)] (\emph{Binary time bound, $\NEXP$-complete.}) Given a
deterministic Turing machine $M$ and $T \in \N$ in binary, does there
exist $w \in \Sigma^*$ and $t \le T$ such that $M$ accepts $w$ in $t$
steps?

\item[(iii)] (\emph{No time bound, $\RE$-complete.}) Given a 
deterministic Turing machine $M$, does there exist $w \in \Sigma^*$
such that $M$ accepts $w$ (in finite time)?
\end{enumerate}
\end{lemma}

\begin{proof}
We treat the three items in turn.

\medskip
\noindent\emph{(i) $\NP$-completeness (unary time bound).}

\textbf{$\NP$-membership.}
Let $(M, T)$ be an instance, with $T$ given in unary. A nondeterministic
polynomial-time verifier guesses $w \in \Sigma^*$ of length at most
$T$ and a time $t \le T$, then simulates $M$ on input $w$ for exactly
$t$ steps. Since $T$ is in unary, the witness $(w, t)$ has polynomial
bit length and the simulation runs in polynomial time. The verifier
accepts iff $M$ is in an accepting state at step $t$.

\textbf{$\NP$-hardness.}
We reduce from $\SAT$. Let $\varphi$ be a Boolean formula in $\CNF$
with $n$ variables. Construct a deterministic Turing machine 
$M_\varphi$ that, on input $w \in \{0,1\}^*$, checks whether 
$|w| = n$ (rejecting otherwise), interprets $w$ as a Boolean 
assignment, and accepts iff $\varphi$ is satisfied under this 
assignment. The machine $M_\varphi$ runs in time $\mathrm{poly}(|\varphi|)$;
let $T$ be a unary encoding of an upper bound on this runtime. Then
$\varphi \in \SAT$ iff $\exists w \in \{0,1\}^n,\, \exists t \le T,\;
M_\varphi$ accepts $w$ in $t$ steps. The reduction is polynomial-time.

\medskip
\noindent\emph{(ii) $\NEXP$-completeness (binary time bound).}

\textbf{$\NEXP$-membership.}
Let $(M, T)$ be an instance with $T$ in binary, of total input size
$n = |M| + |T|$. Since $M$ visits at most $T$ tape positions in $T$
steps, we may restrict to $w \in \Sigma^{\le T}$. A nondeterministic
verifier guesses such a $w$ in $\le T = 2^{O(n)}$ nondeterministic
bits and a time $t \le T$, then simulates $M$ on $w$ for $t$ steps in
deterministic time $\mathrm{poly}(|M|, T) = 2^{O(n)}$. Total
nondeterministic running time is $2^{O(n)}$, hence in $\NEXP$.

\textbf{$\NEXP$-hardness.}
We reduce from $\textsc{Succinct-3SAT}$, the canonical $\NEXP$-complete
problem~\cite{papadimitriou1986note,arora2009}. An instance is a
Boolean circuit $C$ of size $s$ that, on input a $\lceil 2 \log N
\rceil$-bit clause index, outputs three pairs (variable index, sign)
encoding a clause of a $3$-CNF formula $\varphi$ over $N = 2^{|s|}$
variables; the question is whether $\varphi$ is satisfiable. From $C$,
construct a deterministic Turing machine $M_C$ that, on input
$w \in \{0,1\}^N$ interpreted as an assignment, iterates over all
$\mathrm{poly}(N)$ clauses of $\varphi$ (each obtained by simulating
$C$, which runs in time $\mathrm{poly}(s) = \mathrm{poly}(\log N)$)
and accepts iff every clause is satisfied. Then $M_C$ has size 
$\mathrm{poly}(s)$ and runs in time 
$T_0 = \mathrm{poly}(N) \cdot \mathrm{poly}(s)$; let $T = T_0$ in
binary, of size $O(\log N) = \mathrm{poly}(s)$ bits. The equivalence
$\varphi \in \textsc{Succinct-3SAT} \iff \exists w \in \{0,1\}^N,\, 
\exists t \le T,\, M_C$ accepts $w$ in $t$ steps yields the reduction.

\medskip
\noindent\emph{(iii) $\RE$-completeness (no time bound).}

\textbf{$\RE$-membership.}
A semi-decision procedure enumerates pairs $(w, t) \in \Sigma^* \times
\N$ (e.g. by $|w| + t$ increasing) and simulates $M$ on $w$ for $t$
steps; it accepts iff an accepting configuration is reached.

\textbf{$\RE$-hardness.}
We reduce from the halting problem $\HALT_\epsilon = \{ M : M$ halts
on $\epsilon \}$, which is $\RE$-complete~\cite{arora2009}. Given $M$,
let $M'$ be the machine that ignores its input and simulates $M$ on
$\epsilon$. Then $M \in \HALT_\epsilon$ iff $\exists w \in \Sigma^*,\,
M'$ accepts $w$. The map $M \mapsto M'$ is polynomial-time.
\end{proof}

\begin{remark}[Two binary-$T$ variants: $\EXPTIME$ vs.\ $\NEXP$]
\label{rem:exptime-nexp}
With $T$ in binary, two natural variants of the decision problem
above must be distinguished:
\begin{itemize}\setlength{\itemsep}{0pt}
\item \emph{Fixed input.} ``Given $M$, a fixed input $w$, and $T$ in
binary, does $M$ accept $w$ in at most $T$ steps?'' This is
$\EXPTIME$-complete: the deterministic simulation runs in time $T
\le 2^{|T|}$, and hardness follows from the time-hierarchy theorem.
\item \emph{Existential input.} As in Lemma~\ref{lem:tm-complexity}(ii)
above: $w$ is quantified existentially. This is $\NEXP$-complete.
\end{itemize}
The two problems differ in whether $w$ is part of the input or (fixed) quantified existentially. Our reduction in the proof of \Cref{thm:verif-spectrum} encodes $w$ as the real-valued point $x \in \Init$ chosen existentially inside the network's convex
initial region, hence corresponds to the existential variant. This
is why we obtain $\NEXP$, not $\EXPTIME$, for binary $T$.
\end{remark}

By passing to a fixed universal Turing machine, the same statements
hold with $M$ replaced by an instance of a single fixed $U$.

\begin{lemma}[Fixing the machine]
\label{lem:tm-universal}
There exists a fixed deterministic universal Turing machine $U$ such
that the three problems of Lemma~\ref{lem:tm-complexity} retain their
respective complexity classifications when ``$M$ accepts $w$ in $t$
steps'' is replaced by ``$U$ accepts $\langle M, w\rangle$ in at most
$p_U(t, |M|)$ steps'' for some fixed polynomial $p_U$.
\end{lemma}

\begin{proof}
Fix once and for all a deterministic universal Turing machine $U$
with the standard property
(see e.g.~\cite{HMU01,arora2009}) that there
exists a polynomial $p_U$ such that, for every deterministic Turing
machine $M$, every $w \in \Sigma^*$, and every $t \in \N$,
\begin{equation}
\label{eq:utm-overhead}
   M \text{ accepts } w \text{ in } t \text{ steps} 
   \iff
   U  \text{ accepts }\langle M, w\rangle   \text{ in at most } t' \le p_U(|M|, t)
   \text{ steps},
\end{equation}
where $\langle M, w \rangle \in \Sigma^*$ is a standard binary
encoding of the pair, of size $O(|M| + |w|)$.

The map
\(
   \Phi \colon (M, T) \mapsto (M, p_U(|M|, T))
\)
(with $\Phi(M) = M$ in case (iii)) is, by~\eqref{eq:utm-overhead}, a
two-sided reduction between each item of Lemma~\ref{lem:tm-complexity}
and its universal-machine variant. It remains to verify that $\Phi$
is polynomial-time and that the resulting universal-machine variant
falls into the claimed class.

For Lemma~\ref{lem:tm-complexity}(i), $T$ is in unary, so $T' = p_U(|M|, T)$ is computable and
written in unary in polynomial time. For Lemma~\ref{lem:tm-complexity}(ii), $T$ is in binary, so
$|T'| = O(\log p_U(|M|, T)) = \mathrm{poly}(|M|, |T|)$ and $\Phi$ is
polynomial-time. For Lemma~\ref{lem:tm-complexity}(iii), $\Phi$ is the identity on $M$.

In each case, membership in the appropriate class
($\NP$/$\NEXP$/$\RE$) follows by viewing the universal-machine
variant as a special case of Lemma~\ref{lem:tm-complexity} applied to the
single fixed machine $U$ on inputs of the form $\langle M, w \rangle$:
the new time bound is $T' = p_U(T, |M|)$, which has the same
encoding type as $T$ (unary/binary/none, depending on the item);
the size of $\langle M, w \rangle$ is polynomial in $|M|$ and $|w|$.
Hardness follows from Equation~\eqref{eq:utm-overhead} together with
the hardness part of the corresponding item of
Lemma~\ref{lem:tm-complexity}.

\end{proof}

\subsection{Proof of \Cref{thm:verif-spectrum}}
\label{app:verif-spectrum-proof}

We now assemble the ingredients to give a formal proof of 
\Cref{thm:verif-spectrum}, restating the result for convenience.

\begin{theorem*}[\Cref{thm:verif-spectrum}, restated]
The bounded-time recurrent reachability problem 
(Definition~\ref{def:bounded-time-reach}) for width-$4$ $\ReLU$ networks is:
\begin{enumerate}\setlength{\itemsep}{0pt}
   \item[\textnormal{(i)}] $\NP$-complete when $T$ is encoded in unary;
   \item[\textnormal{(ii)}] $\NEXP$-complete when $T$ is encoded in binary;
   \item[\textnormal{(iii)}] $\RE$-complete (undecidable) when $T$ is unbounded.
\end{enumerate}
\end{theorem*}

\paragraph{Reduction recipe (uniform across (i)--(iii)).}
For each item, we reduce a Turing-machine acceptance problem to
width-$4$ recurrent reachability. The reduction is identical across
items; only the source problem and the encoding of $T$ vary.

Let $M$ be a deterministic Turing machine and $T \in \N$ a time
bound (unary, binary, or absent). By \Cref{thm:main}, applied to 
$M$, we obtain a width-$4$ $\ReLU$ network 
$N_M \colon [0,1]^4 \to [0,1]^4$ of depth $D_M = \mathrm{poly}(|M|)$,
constructible in polynomial time, with the simulation property 
(part~(i)) and the online certificate property (part~(ii)). Recall
from \Cref{app:width3} that the second coordinate $x_2$ co-encodes
the state and right tape: a state-prefix $\enc_M(q) \in 
\{0,1,\dots,4^{|q|_4} - 1\} \cdot 4^{-|q|_4}$ of length 
$|q|_4 = \lceil \log_4 |Q| \rceil$ is followed by a 
$\onethreefouradic$-encoded right tape.

Define the convex regions
\begin{align*}
   \Init &:= \bigl\{ x \in [0,1]^4 \;:\;
        \enc_M(q_0) \le 4^{|q|_4}\, x_2 \le \enc_M(q_0) + 1,\;
        x_3 = 0,\; x_4 = 0 \bigr\}, \\
   \Acc &:= \bigl\{ x \in [0,1]^4 \;:\;
        \enc_M(q_F) \le 4^{|q|_4}\, x_2 \le \enc_M(q_F) + 1,\;
        x_4 \le 0 \bigr\}.
\end{align*}
Both regions are specified by axis-aligned linear inequalities with
rational coefficients of bit complexity $O(|q|_4) = O(\log |M|)$, 
hence convex and polynomial-time constructible from $M$.

The reduction maps the Turing-machine instance $(M, T)$ (or just 
$M$ for case (iii)) to the reachability instance 
$(N_M, \Init, \Acc, T \cdot C_M)$, where $C_M$ is the per-step
iteration count of \Cref{thm:main}.

\paragraph{Correctness of the recipe.}
We claim:
\begin{equation}
\label{eq:recipe-correctness}
\begin{aligned}
   &\text{There exist } x \in \Init \text{ and } t \le T \cdot C_M
     \text{ with } N_M^{[t]}(x) \in \Acc \\
   &\quad\iff\quad
   \exists\, \tau \in \{0,1\}^\omega,\; \exists\, t' \le T,\;
     M \text{ accepts } \tau \text{ in } t' \text{ steps}.
\end{aligned}
\end{equation}

\emph{($\Rightarrow$)} Suppose $x \in \Init$ and $N_M^{[t]}(x) \in
\Acc$ for some $t \le T \cdot C_M$. We have $x_4(0) = 0$ and 
$\pi_4(N_M^{[t]}(x)) \le 0$ (since $N_M^{[t]}(x) \in \Acc$). 
By monotonicity of $x_4$ along the trajectory
(Lemma~\ref{lem:invariant}(a)), this forces
$\pi_4(N_M^{[k]}(x)) = 0$ for every $k \le t$. By
items (b)--(c) of Lemma~\ref{lem:invariant}, every step $k \le t$ is
locally valid: the leading digit of $x_1^{(k)}$ and of the
tape-suffix of $x_2^{(k)}$ lies in $\{1, 3\}$, and the state-prefix
of $x_2^{(k)}$ lies in $\enc_M(Q)$. Since $N_M$ acts by affine
maps that are uniform on leading-digit cylinders, the state prefixes
of its first $t$ iterates depend only on the digits visited during
the first $t$ iterations. Let $x'$ be obtained from $x$ by setting
every unvisited digit of $x_1$ and of the tape-suffix of $x_2$ to
$1$. Then $x'$ encodes a configuration $c_0(\tau)$ with
$\tau \in \{0,1\}^\omega$, and $N_M^{[k]}(x)$ and $N_M^{[k]}(x')$
have the same state prefix for every $k \le t$. By
\Cref{thm:main}(i), $N_M$ faithfully simulates $M$ on the
infinite binary tape $\tau$ encoded by $x'$, and its trajectory
reaches a state-encoded $q_F$ at iteration $\le t$. Hence $M$ accepts $\tau$ in some $t' \le t / C_M \le T$
steps.

\emph{($\Leftarrow$)} Conversely, suppose $M$ accepts some 
$\tau \in \{0,1\}^\omega$ in $t' \le T$ steps. Set 
$x = \enc_M(c_0(\tau)) \in \Init$. By \Cref{thm:main}(i),
$N_M^{[C_M\, t']}(x) = \enc_M(c_{t'}(\tau))$, whose state-prefix is
$\enc_M(q_F)$ and whose fourth coordinate is $0$ (by 
Lemma~\ref{lem:invariant}(b), since the trajectory remains within 
encoded configurations). Hence $N_M^{[C_M\, t']}(x) \in \Acc$ with 
$C_M\, t' \le T \cdot C_M$.

\paragraph{Polynomial-time computability.}
The reduction $(M, T) \mapsto (N_M, \Init, \Acc, T \cdot C_M)$ is
polynomial-time: $N_M$ is constructed in $\mathrm{poly}(|M|)$ by 
\Cref{thm:main}; $\Init$ and $\Acc$ are described by $O(|q|_4) = 
O(\log |M|)$ rational coefficients; and the new time bound $T \cdot 
C_M$ has size $|T| + \mathrm{poly}(|M|)$, which preserves the 
encoding (unary stays unary, binary stays binary).

\paragraph{Hardness of (i)--(iii).}
Apply the recipe with the source problems of Lemma~\ref{lem:tm-complexity}:
\begin{itemize}\setlength{\itemsep}{0pt}
\item (i) Source: Lemma~\ref{lem:tm-complexity}(i), $\NP$-complete with 
unary $T$. The recipe yields a polynomial-time reduction to 
width-$4$ reachability with unary time bound $T \cdot C_M$.

\item (ii) Source: Lemma~\ref{lem:tm-complexity}(ii), $\NEXP$-complete 
with binary $T$. The recipe yields a polynomial-time reduction 
preserving the binary encoding of the time bound.

\item (iii) Source: Lemma~\ref{lem:tm-complexity}(iii), $\RE$-complete 
with no time bound. The recipe yields a polynomial-time reduction
to unbounded reachability.
\end{itemize}

\paragraph{Membership in (i): $\NP$.}
Given $(M, T)$ with $T$ in unary, build $(N_M, \Init, \Acc, T \cdot 
C_M)$ in polynomial time. By Lemma~\ref{lem:np-witness} applied with 
parameters $d = 4$, $w = 4$, $L = D_M = \mathrm{poly}(|M|)$, time
bound $T \cdot C_M = \mathrm{poly}(|M|, T)$, weight bit complexity
$b = \mathrm{poly}(|M|)$, and region bit complexity $b' = O(\log
|M|)$, any reachability witness can be taken rational of bit 
complexity $\mathrm{poly}(|M|, T)$. A nondeterministic verifier 
guesses such a rational $x \in \Init$, simulates $N_M$ exactly for
$T \cdot C_M$ iterations, and checks at each step whether the 
trajectory enters $\Acc$. Total time is $\mathrm{poly}(|M|, T)$, 
hence $\NP$-membership.

\paragraph{Membership in (ii): $\NEXP$.}
Given $(M, T)$ with $T$ in binary, build $(N_M, \Init, \Acc, T 
\cdot C_M)$ as before. Apply Lemma~\ref{lem:np-witness} with the same
parameters except $T \cdot C_M$ now binary, of bit length 
$|T| + O(\log |M|)$. The lemma yields a rational witness of bit
complexity $\mathrm{poly}(|M|, T)$, which is exponential in the
input size $|M| + |T|$. A nondeterministic verifier guesses such a
witness in $\mathrm{poly}(|M|, T) = 2^{\mathrm{poly}(|M|, |T|)}$ 
nondeterministic time, then simulates $N_M$ for $T \cdot C_M = 
2^{O(|T|)} \cdot \mathrm{poly}(|M|)$ steps in deterministic 
exponential time. Total nondeterministic running time is 
$2^{\mathrm{poly}(|M|, |T|)}$, hence $\NEXP$-membership.

\paragraph{Membership in (iii): $\RE$.}
A semi-decision procedure simulates $N_M$ from each rational point 
of $\Init$ enumerated in some computable order, for increasing 
numbers of iterations, accepting if the trajectory ever enters 
$\Acc$. By Lemma~\ref{lem:np-witness}, if the instance is positive then
some rational $x \in \Init$ of finite bit complexity reaches $\Acc$
in finite time, so the procedure halts and accepts on positive 
instances. Hence the problem is in $\RE$.

\medskip
Combining hardness and membership in each case proves
\Cref{thm:verif-spectrum}.

\section{Robust hardness transfer from \texorpdfstring{$\ReLU$}{ReLU}
to smooth activations}
\label{app:robust-transfer}

This appendix proves \Cref{thm:rho} by transferring $\NP$-hardness
from $\ReLU$ networks at width $4$ to $\rho$-networks at width $6$. We
work with the \emph{register} form of $\ReLU$ networks, in which each
hidden layer updates a single coordinate.

The key technical ingredient, used to construct identity-preserving
gadgets and approximate $\ReLU$ from $\rho$, is the local-identity
approximation of~\citet[Lemma~4.1]{kidger2020universal}.

\begin{proposition}[Local approximation of the identity, {\citealp[Lemma 4.1]{kidger2020universal}}]
\label{kidger41}
Let $\rho \colon \R \to \R$ be continuous and continuously
differentiable at at least one point with nonzero derivative there.
Let $L \subseteq \R$ be compact. Then a single enhanced neuron with
activation $\rho$ uniformly approximates the identity function
$\mathrm{Id} \colon \R \to \R$ on $L$, with arbitrarily small error.
\end{proposition}

\begin{proof}[Proof sketch (after~\citealp{kidger2020universal})]
By hypothesis there exist $\alpha \in \R$ where $\rho$ is
continuously differentiable with $\rho'(\alpha) \ne 0$. For
$h \in \R \setminus\{0\}$ set $\phi_h(x) = h x + \alpha$ and
$\psi_h(x) = (x - \rho(\alpha))/(h \rho'(\alpha))$. Then
$\iota_h = \psi_h \circ \rho \circ \phi_h$ is realized by a single
enhanced neuron, and $\iota_h(x) \to x$ uniformly on any compact $L$
as $h \to 0$, by the mean-value theorem applied to $\rho$ on 
$[\alpha, \alpha + h x]$.
\end{proof}

\subsection{Register networks}
\label{app:robust-register}\label{app:register}

\begin{definition}[$\ReLU$-register network, restated from \Cref{def:relu-register-net}]
A $\ReLU$-register network $F \colon \R^d \to \R^d$ of register width 
$k$ and depth $L$ is a composition
\[
   F = A_{\mathrm{out}} \circ R_L \circ \cdots \circ R_1 \circ A_{\mathrm{in}}
   \colon \R^d \to \R^d,
\]
where $A_{\mathrm{in}} \colon \R^d \to \R^k$ and 
$A_{\mathrm{out}} \colon \R^k \to \R^d$ are affine, and each
register-layer $R_\ell \colon \R^k \to \R^k$ is of the form: there 
exist an index $j_\ell \in \{1, \dots, k\}$ and parameters 
$a_\ell \in \R^k$, $b_\ell \in \R$ such that 
$(R_\ell(x))_i = x_i$ for $i \neq j_\ell$ and 
$(R_\ell(x))_{j_\ell} = \ReLU(a_\ell \cdot x + b_\ell)$.
\end{definition}

\begin{remark}[Native register form of $N_M$]
\label{rem:nm-register}
The network $N_M$ of \Cref{thm:main} is, by its construction, in 
register form of register width $4$:
\begin{itemize}\setlength{\itemsep}{0pt}
\item the width-$3$ simulator $g_M$ of \Cref{thm:simulation-optimize}
is in register form by the construction of~\cite{MFCS2025};
\item the certificate update 
$x_4 \leftarrow \max\{x_4, \Cert(x_1, x_2)\}$ modifies a single 
coordinate, computed in width $2$ via 
$\max\{a, b\} = a + \ReLU(b - a)$;
\item the unmodified coordinates are propagated unchanged.
\end{itemize}
In particular, $N_M$ can be passed directly to the $\rho$-emulation
of Lemma~\ref{lem:rho-emulation-network} below without any preliminary
register-form conversion.
\end{remark}

\subsection{Approximation of $\ReLU$ register networks by smooth networks}
\label{app:robust-approximation}\label{app:rho-emulation}

We approximate $\ReLU$-register networks by $\rho$-networks in two
steps: first one register-layer at a time
(Lemma~\ref{lem:relu-register-to-rho}), then the entire network by 
telescoping (Lemma~\ref{lem:rho-emulation-network}).

This is precisely Lemma~\ref{lem:rho-emulation} of the main body, which we restate and prove here.

\begin{lemma}[Approximating one $\ReLU$ register-layer at width $k+2$]
\label{lem:relu-register-to-rho}
Let $\rho \colon \R \to \R$ be continuous, non-polynomial, and $C^1$
at some point with non-zero derivative. Fix $k \ge 1$, a compact 
set $K \subseteq \R^k$, an index $j \in \{1, \dots, k\}$, and 
parameters $a \in \R^k$, $b \in \R$. Consider the $\ReLU$ 
register-layer $F \colon \R^k \to \R^k$ with $F_i(x) = x_i$ for 
$i \neq j$ and $F_j(x) = \ReLU(a \cdot x + b)$. Then for every 
$\varepsilon > 0$ there exists a $\rho$-network 
$\widetilde F \in \Netnotation{\rho}{k}{k}{k+2}$, constructible in
time $\mathrm{poly}(k, \log(1/\varepsilon), \mathrm{size}(a, b))$, 
such that
\[
   \sup_{x \in K} \|\widetilde F(x) - F(x)\|_\infty \le \varepsilon.
\]
\end{lemma}

\begin{proof}
The strategy is to dedicate $k$ of the $k + 2$ neurons to 
approximating the identity on the $k$ register coordinates, and the
remaining $2$ neurons to approximating the ReLU operation 
$z \mapsto \ReLU(a \cdot x + b)$. Coordinates are then assembled 
via the output affine map.

\smallskip
\noindent\emph{Step 1: ingredients.}
By hypothesis on $\rho$ and Proposition~\ref{kidger41}, for every compact 
$L \subseteq \R$ and every $\eta > 0$ there exists a single 
enhanced $\rho$-neuron computing a function 
$\iota_\eta \colon \R \to \R$ with 
$\sup_{z \in L} |\iota_\eta(z) - z| \le \eta$.

By the effective ReLU emulation result of~\citet{zhang2024deep} 
(see also~\citealp[Theorem~4.3]{kidger2020universal} for existence),
for activations satisfying (H1)--(H3) and every compact 
$L' \subseteq \R$ and every $\eta > 0$, there exists a $\rho$-network 
$\mathcal{R}_\eta \colon \R \to \R$ of width $2$ and constant depth, 
constructible in time $\mathrm{poly}(\log(1/\eta), \log B)$ where 
$B = \sup_{z \in L'} |z|$, such that 
$\sup_{z \in L'} |\mathcal{R}_\eta(z) - \ReLU(z)| \le \eta$.

\smallskip
\noindent\emph{Step 2: compact sets used.}
Let $C_a := \sup_{x \in K} |a \cdot x + b|$, finite by compactness
of $K$, and set $L' := [-C_a, C_a]$. Let $L \subseteq \R$ be a 
compact set containing every coordinate value $x_i$ for $x \in K$,
$i \in \{1, \dots, k\}$.

\smallskip
\noindent\emph{Step 3: construction of $\widetilde F$.}
Set $\eta := \varepsilon$ and define 
$\widetilde F \in \Netnotation{\rho}{k}{k}{k+2}$ as follows. The 
hidden layer has $k + 2$ neurons:
\begin{itemize}\setlength{\itemsep}{0pt}
\item Neurons $1, \dots, k$: each neuron $i$ takes input $x_i$ via
the input affine map and applies the identity-emulating gadget 
$\iota_\eta$ on $L$. By Step~1, neuron $i$ outputs 
$\iota_\eta(x_i)$ with $|\iota_\eta(x_i) - x_i| \le \eta$ for 
$x \in K$.

\item Neurons $k + 1$ and $k + 2$: jointly compute the ReLU gadget
$\mathcal{R}_\eta$ applied to the pre-activation $a \cdot x + b$,
formed via the input affine map. By Step~1, the combined output 
approximates $\ReLU(a \cdot x + b)$ within $\eta$ on $L'$.
\end{itemize}
The output affine map assembles the $k$ output coordinates: for 
$i \neq j$, set $\widetilde F_i(x) := \iota_\eta(x_i)$; for 
$i = j$, set $\widetilde F_j(x) := \mathcal{R}_\eta(a \cdot x + b)$.

\smallskip
\noindent\emph{Step 4: error bound.}
For every $x \in K$ and every coordinate $i$:
\begin{itemize}\setlength{\itemsep}{0pt}
\item if $i \neq j$: $|\widetilde F_i(x) - F_i(x)| = |\iota_\eta(x_i) - x_i| \le \eta = \varepsilon$;
\item if $i = j$: $|\widetilde F_j(x) - F_j(x)| = |\mathcal{R}_\eta(a \cdot x + b) - \ReLU(a \cdot x + b)| \le \eta = \varepsilon$.
\end{itemize}
Hence $\|\widetilde F(x) - F(x)\|_\infty \le \varepsilon$ uniformly
on $K$.

\smallskip
\noindent\emph{Step 5: constructibility.}
Both $\iota_\eta$ and $\mathcal{R}_\eta$ are constructed by the 
proofs of Proposition~\ref{kidger41} and Theorem~4.3 of~\cite{kidger2020universal}
in time $\mathrm{poly}(\log(1/\eta), \mathrm{size}(a, b))$.
\end{proof}

We now lift the per-layer approximation to a network-level statement.
This is precisely Lemma~\ref{lem:rho-emulation-network} of the main body,
which we restate and prove here.

\begin{lemma}[Lemma~\ref{lem:rho-emulation-network}, restated]
\label{lem:rho-emulation-network-restated}
Let $\rho \colon \R \to \R$ be continuous, non-polynomial, and $C^1$
at some point with non-zero derivative. Let $F \colon \R^d \to \R^d$
be a $\ReLU$-register network of register width $k$ and depth $L$,
let $K \subseteq \R^d$ be compact, and let $\varepsilon > 0$. Then 
there exists a $\rho$-network 
$\widetilde F \in \Netnotation{\rho}{d}{d}{k+2}$ of depth $O(L)$, 
constructible in time $\mathrm{poly}(|F|, \log(1/\varepsilon))$, 
such that
\[
   \sup_{x \in K} \|\widetilde F(x) - F(x)\|_\infty \le \varepsilon.
\]
\end{lemma}

\begin{proof}
Write $F = A_{\mathrm{out}} \circ R_L \circ \cdots \circ R_1 \circ
A_{\mathrm{in}}$. Let $L_\ell := \max\{1, \|a_\ell\|_1\}$ be the 
Lipschitz constant of $R_\ell$ in $\|\cdot\|_\infty$ (since 
$\ReLU$ is $1$-Lipschitz and $|a_\ell \cdot u| \le \|a_\ell\|_1
\|u\|_\infty$); let $L_{\mathrm{out}}$ be the Lipschitz constant 
of $A_{\mathrm{out}}$.

\smallskip
\noindent\emph{Step 1: compact propagation with buffer.}
Set $K_0 := A_{\mathrm{in}}(K)$ and inductively 
$K_\ell := R_\ell(K_{\ell-1})$, all compact in $\R^k$. Let 
$\eta := \varepsilon / L_{\mathrm{out}}$ and define the enlarged
compacts $K_\ell^+ := K_\ell + \overline{B_\eta}$ (closed 
$\eta$-neighborhood in $\|\cdot\|_\infty$).

\smallskip
\noindent\emph{Step 2: per-layer approximation.}
Apply Lemma~\ref{lem:relu-register-to-rho} on $K_{\ell - 1}^+$ with 
target precision
\[
   \delta_\ell := \frac{\varepsilon}{L_{\mathrm{out}} \cdot (L + 1)
                       \cdot \prod_{t = \ell + 1}^{L} L_t},
\]
producing a $\rho$-block 
$\widetilde R_\ell \in \Netnotation{\rho}{k}{k}{k+2}$ such that
\[
   \sup_{u \in K_{\ell - 1}^+} \|\widetilde R_\ell(u) - R_\ell(u)\|_\infty
   \;\le\; \delta_\ell.
\]
The block is constructible in time 
$\mathrm{poly}(k, \log(1/\delta_\ell), \mathrm{size}(a_\ell, b_\ell))
= \mathrm{poly}(|F|, \log(1/\varepsilon))$.

\smallskip
\noindent\emph{Step 3: composed network.}
Define $\widetilde F := A_{\mathrm{out}} \circ \widetilde R_L \circ
\cdots \circ \widetilde R_1 \circ A_{\mathrm{in}}$. Each $\widetilde
R_\ell$ has width $k + 2$ and constant depth, so $\widetilde F$ has
width $k + 2$ and depth $O(L)$.

\smallskip
\noindent\emph{Step 4: error telescoping.}
For $x \in K$, let 
$\xi_\ell(x) := R_\ell \circ \cdots \circ R_1 \circ A_{\mathrm{in}}(x)
\in K_\ell$ and 
$\widetilde \xi_\ell(x) := \widetilde R_\ell \circ \cdots \circ
\widetilde R_1 \circ A_{\mathrm{in}}(x)$, with 
$E_\ell(x) := \|\widetilde \xi_\ell(x) - \xi_\ell(x)\|_\infty$ and 
$E_0 \equiv 0$.

We claim by induction that $\widetilde \xi_\ell(x) \in K_\ell^+$ 
and 
$E_\ell(x) \le \sum_{j=1}^{\ell} \delta_j \prod_{t = j + 1}^{\ell} L_t$
for every $x \in K$ and $\ell \le L$. The base case $\ell = 0$ is
immediate. Assume both claims for $\ell - 1$. Then 
$\widetilde \xi_{\ell - 1}(x) \in K_{\ell - 1}^+$, so the 
approximation guarantee for $\widetilde R_\ell$ applies at this 
point, giving
\begin{align*}
   E_\ell(x)
   &\le \|\widetilde R_\ell(\widetilde \xi_{\ell - 1}) 
         - R_\ell(\widetilde \xi_{\ell - 1})\|_\infty 
       + \|R_\ell(\widetilde \xi_{\ell - 1}) 
         - R_\ell(\xi_{\ell - 1})\|_\infty \\
   &\le \delta_\ell + L_\ell \cdot E_{\ell - 1}(x).
\end{align*}
Solving the recurrence yields 
$E_\ell(x) \le \sum_{j=1}^{\ell} \delta_j \prod_{t = j+1}^{\ell} L_t$.
By the choice of $\delta_j$, each summand satisfies
\[
   \delta_j \prod_{t = j + 1}^{\ell} L_t
   = \frac{\varepsilon}{L_{\mathrm{out}}\,(L + 1)} \cdot 
     \frac{\prod_{t = j + 1}^{\ell} L_t}{\prod_{t = j + 1}^{L} L_t}
   \le \frac{\varepsilon}{L_{\mathrm{out}}\,(L + 1)},
\]
so $E_\ell(x) \le L \varepsilon / (L_{\mathrm{out}} (L + 1))
\le \varepsilon / L_{\mathrm{out}} = \eta$ uniformly. This shows in
particular $\widetilde \xi_\ell(x) \in K_\ell^+$, validating the
induction.

\smallskip
\noindent\emph{Step 5: final bound.}
Apply $A_{\mathrm{out}}$:
\[
   \|\widetilde F(x) - F(x)\|_\infty
   = \|A_{\mathrm{out}}(\widetilde \xi_L) - A_{\mathrm{out}}(\xi_L)\|_\infty
   \le L_{\mathrm{out}} \cdot E_L(x) < \varepsilon
\]
uniformly on $K$.

\smallskip
\noindent\emph{Constructibility.}
The constants $L_\ell$ and $L_{\mathrm{out}}$ are computable from 
the affine parameters of $F$ in linear time. Each $\delta_\ell$ has 
$\log(1/\delta_\ell) = \mathrm{poly}(|F|, \log(1/\varepsilon))$ 
bits. By Step~2, each $\widetilde R_\ell$ is constructible in time
$\mathrm{poly}(|F|, \log(1/\varepsilon))$, and the composition is
linear in $L$.
\end{proof}

\subsection{Robust hardness transfer}
\label{app:robust-transfer-proof}

\begin{theorem}[Robust hardness transfer $\ReLU \to \rho$]
\label{thm:robust-transfer}
Fix $d \ge 1$ and $k \ge d$, and let $\rho$ satisfy hypotheses 
(H1)--(H3) of \Cref{thm:rho}. If $\delta$-robust verification 
(resp.\ $\delta$-robust bounded-time reachability) is $\NP$-hard 
for width-$k$ $\ReLU$-register networks in dimension $d$, then the
corresponding problem is $\NP$-hard for width-$(k + 2)$ 
$\rho$-networks.
\end{theorem}

\begin{proof}
We give a polynomial-time many-one promise-preserving reduction.
Given an instance $(N, \Init, R, \delta)$ of $\delta$-verification 
for a $\ReLU$-register network $N$ of width $k$, apply 
Lemma~\ref{lem:rho-emulation-network-restated} on the compact 
$K := \Init$ with target precision $\varepsilon := \delta / 4$ to 
obtain a width-$(k + 2)$ $\rho$-network $\widetilde N$ satisfying 
$\sup_{x \in K} \|\widetilde N(x) - N(x)\|_\infty \le \delta / 4$.
Output the instance 
$(\widetilde N, \Init, R^{+\delta}, \delta / 4)$, where $R^{+\delta}$
denotes the closed $\delta$-neighborhood of $R$.

\emph{YES preservation.} If there exists $x \in \Init$ with 
$N(x) \in R^{+\delta}$, then 
$\dist(\widetilde N(x), R) \le \dist(N(x), R) + \delta / 4 \le 
\delta + \delta / 4$, so 
$\widetilde N(x) \in R^{+(\delta + \delta/4)} \subseteq
(R^{+\delta})^{+\delta/4}$: the output is YES at the new promise
gap $\delta / 4$.

\emph{NO preservation.} If $\dist(N(x), R) > 2\delta$ for all 
$x \in \Init$, then $\dist(N(x), R^{+\delta}) > \delta$, hence
$\dist(\widetilde N(x), R^{+\delta}) \ge \dist(N(x), R^{+\delta}) -
\delta / 4 > 3 \delta / 4 = 2 \cdot (\delta / 4) + \delta / 4 > 
2 \cdot (\delta / 4)$, so the output is NO.

For bounded-time reachability with unary $T$, apply the same 
reduction with $K = [0, 1]^d$ and 
$\varepsilon := \delta / (4 \cdot S_T)$ where 
$S_T := \sum_{i = 0}^{T - 1} L^i$ and $L$ is a Lipschitz constant 
of $N$ on $K$; the standard error-recurrence telescoping 
(cf. Step~4 of Lemma~\ref{lem:rho-emulation-network-restated}) controls
the iterated error within $\delta / 4$ across $T$ iterations.

The reduction is polynomial-time by the constructibility bound of
Lemma~\ref{lem:rho-emulation-network-restated}.
\end{proof}

\begin{theorem}[$\NP$-membership for robust verification of smooth networks]
\label{thm:rho-np-membership}
Fix $d \ge 1$, and let $\rho$ satisfy hypotheses (H1)--(H3) of 
\Cref{thm:rho}. Then $\delta$-robust verification (resp.\ robust
bounded-time reachability with unary $T$) for $\rho$-networks of
dimension $d$ belongs to $\NP$.
\end{theorem}

\begin{proof}
A nondeterministic verifier guesses a rational $x \in \Init$ with 
polynomially many bits (justified below) and evaluates 
$\widetilde y := N(x)$ to precision $\delta / 4$ using 
polynomial-time evaluation of $\rho$ (by (H3)) and exact rational
arithmetic for the affine layers.

\emph{Witness bit-complexity.} 
By hypothesis (H3), $\rho$ admits a polynomial-time-computable 
Lipschitz constant $L_\rho$ on every bounded interval. Combined 
with the Lipschitz constants of the affine layers, $N$ is Lipschitz
with constant $L_N = \mathrm{poly}(\text{input})$ on $K = \Init$. 
By Lemma~\ref{lem:relu-register-to-rho} applied at precision $\delta /4$,
there is a $\ReLU$-equivalent $N^{\ReLU}$ with 
$\sup_{x \in K} \|N(x) - N^{\ReLU}(x)\|_\infty \le \delta / 4$. 
Apply Lemma~\ref{lem:np-witness} to $N^{\ReLU}$: any feasible real 
witness $x \in \Init$ for the $\ReLU$ instance can be replaced by a
rational witness of bit complexity $\mathrm{poly}(\text{input})$.
The promise gap of $\delta$ vs. $2 \delta$ tolerates the cumulative
error $2 \cdot \delta / 4 = \delta / 2$, ensuring that the rational
witness preserves the YES/NO classification.

\emph{Verification.} The verifier evaluates $\widetilde y$ to 
precision $\delta / 4$. Under the $\delta$ vs.\ $2 \delta$ promise,
$\dist(\widetilde y, R) \le \delta + \delta / 4$ in the YES case 
and $\dist(\widetilde y, R) \ge 2\delta - \delta / 4 > 
\delta + \delta / 4$ in the NO case, so the approximate output 
unambiguously certifies the correct case.
\end{proof}

\medskip
\noindent\textbf{Conclusion.}
Combining \Cref{thm:robust-transfer} with $\NP$-hardness of robust
$\ReLU$ verification at width $4$ (the unary-$T$ case of 
\Cref{thm:verif-spectrum}, equivalently Corollary~\ref{cor:approx}), and 
\Cref{thm:rho-np-membership} for membership, yields \Cref{thm:rho}.

\paragraph{On binary $T$ for $\rho$.}
\label{sec:rho-binary}
The argument above relies on $T$ unary so that the cumulative
approximation error $T \cdot L_\rho^T \cdot \varepsilon_{\mathrm{step}}$
remains controllable in polynomial time. With $T$ binary, the same
calculation requires $\varepsilon_{\mathrm{step}} \le \delta / (4 \cdot 2^{|T|} \cdot L_\rho^{2^{|T|}})$,
which has $\Omega(L_\rho^{2^{|T|}})$ bits of precision, doubly
exponential in the input. Whether $\delta$-robust bounded-time
reachability of $\rho$-networks at width $6$ admits a $\NEXP$-style
hardness statement at binary $T$ remains open.

\paragraph{Approximation hardness for $\rho$-networks.}
The approximation-hardness of Corollary \ref{cor:approx} extends to $\rho$-networks
at width $6$ via the same reduction: under (H1)-(H3) there
exists $\eta > 0$ such that approximating the worst-case output of a
width-$6$ $\rho$-network within additive error $\eta$ is $\NP$-hard
(unary $T$). The argument is identical to the proof above, using that
the $\Omega(1)$ promise gap of Corollary~\ref{cor:approx} is preserved up to a
constant by the $\rho$-emulation. Combined with  Corollary  \ref{cor:approx}, this
yields a unified intractability picture for $\ReLU$ and all common
smooth activations at constant width.

\subsection{Proof of Corollary \ref{cor:np-verif-width4}}
\label{app:cor-np-verif-width4}

\begin{corollary*}[Corollary \ref{cor:np-verif-width4}, restated]
Verification of feedforward width-$4$ $\ReLU$ networks on convex
$\Init, \Acc \subseteq [0,1]^4$ is $\NP$-complete.
\end{corollary*}

\begin{proof}
\emph{$\NP$-membership.} 
Apply \Cref{lem:np-witness} with $T = 1$ (no iteration): a 
nondeterministic verifier guesses a rational $x \in \Init$ of 
polynomial bit complexity, evaluates $N(x)$ exactly, and checks
$N(x) \in \Acc$.

\emph{$\NP$-hardness.}
We reduce from \Cref{thm:verif-spectrum}(i). Given a $T$-bounded 
reachability instance $(N_M, \Init, \Acc, T)$ with $T$ in unary, 
where $N_M$ and $\Init, \Acc$ are as constructed in 
\Cref{app:verif-spectrum-proof}, output the unrolled feedforward 
network $N := N_M^{[T \cdot C_M]}$ together with $\Init$ and $\Acc$
unchanged. The unrolled $N$ has width $4$, depth $T \cdot C_M \cdot 
D_M = \mathrm{poly}(|M|, T)$, and is polynomial-time constructible
since $T$ is unary.

It remains to verify the equivalence
\(
   \exists x \in \Init : N(x) \in \Acc
   \iff
   \exists x \in \Init, \exists t \le T \cdot C_M : N_M^{[t]}(x) 
   \in \Acc.
\)
The reverse direction is immediate by absorption: in the 
construction of \Cref{app:verif-spectrum-proof}, $\Acc$ corresponds
to ``state $q_F$ reached and $x_4 = 0$''. Since $q_F$ is the unique
accepting (and thus terminal) state of $M$, once the trajectory 
enters $\Acc$, $N_M$ acts as the identity on $\Acc$ at every 
subsequent iteration, so $N_M^{[T \cdot C_M]}(x) \in \Acc$ as well.
The forward direction is trivial (take $t = T \cdot C_M$).
\end{proof}

\section{Statements for various activation functions}
\label{app:activations}

The hypotheses of \Cref{thm:rho} ($\rho$ continuous, non-polynomial,
polynomial-time computable, $C^1$ at some point with non-zero derivative)
are mild and cover every smooth activation in modern use. The following
table summarizes which standard activations satisfy the assumptions and
inherit the consequent complexity statements at width $6$.

 \begin{table}[h]
\centering
\caption{Coverage of \Cref{thm:rho} on standard activations.}
\label{tab:activations}
\small
\begin{tabular}{lcl}
\toprule
Activation $\rho$ & Conditions of \Cref{thm:rho} & Notes \\
\midrule
$\tanh$                             & yes & smooth on $\R$ \\
sigmoid $\sigma(x)=(1+e^{-x})^{-1}$ & yes & smooth on $\R$ \\
GELU $x \cdot \Phi(x)$              & yes & smooth on $\R$ \\
Swish $x \cdot \sigma(\beta x)$     & yes & smooth on $\R$, any $\beta > 0$ \\
ELU                                 & yes & \makecell[l]{$C^1$ except at $0$;\\ assumption holds at any other point} \\
softplus $\log(1 + e^x)$            & yes & smooth on $\R$ \\
$\arctan$                           & yes & smooth on $\R$ \\
Gaussian $e^{-x^2}$                 & yes & smooth on $\R$ \\
\bottomrule
\end{tabular}
\end{table}

For \emph{exact} verification (without robustness), much stronger lower
bounds hold for several of these activations. The following result of
\citet{wurm2,wurm2023complexity} shows that exact verification is
$\exists\R$-hard for a broad class of activations, including those
permitting transcendental constraints.

\begin{theorem}[\citealt{wurm2,wurm2023complexity}]
\label{thm:exists-r-hardness}
With identity nodes allowed:
\begin{enumerate}
\item Verification is $\exists\R$-complete for any non-linear polynomial
$\rho$, and $\exists\R$-hard whenever $\rho$ coincides with a non-linear
polynomial on an interval.
\item Verification is $\exists\R$-hard for any activation $\rho$ that
allows pp-defining a function coinciding with the exponential on an
interval (in particular, the exponential, ELU, $e^{-|x|}$, the Gaussian
$e^{-x^2}$, $\arctan$, and the Gallant--White cosine-smasher).
\end{enumerate}
\end{theorem}

These results explain why we restrict our analysis to the \emph{robust}
variant: exact verification falls outside the standard polynomial
hierarchy, while robust verification recovers $\NP$-completeness
(\Cref{thm:rho}).

\section{Neural ODEs at bounded dimension}
\label{app:neural-ode}

Our results are stated for discrete iterated networks. In this section
we transfer them to Neural ODEs~\citep{chen2018neural} via the natural
correspondence established by~\citet{yifei2024vanilla}, obtaining a
\emph{Lipschitz piecewise-affine} vector field of dimension $5$ whose
splitting-method-discretized flow simulates an arbitrary Turing
machine. Verification hardness transfers automatically. The pure
continuous-time question remains open with a precise technical
obstruction.

\subsection{The Duan et al.\ correspondence}
\label{sec:neural-ode-duan}
 
\citet{yifei2024vanilla} establish that vanilla feedforward
networks, not only residual ones, are exact numerical discretizations
of (non-autonomous) dynamical systems. Their Theorem~1 shows that for
every flow map $\phi^\tau$ of a Lipschitz ODE on a compact set, there
exists a vanilla leaky-ReLU FNN of width $d$ approximating it; the
construction proceeds via splitting of an associated neural ODE. The
mechanism implicit in their construction also yields the converse
direction we need: every leaky-ReLU FNN of width exactly $d$ with
non-singular layers \emph{is} the splitting-method iteration of an
explicit non-autonomous piecewise-affine vector field. We state and
use this converse direction below.
 
\begin{lemma}[Splitting interpretation, after~\citet{yifei2024vanilla}]
\label{thm:duan-correspondence}
Let $\sigma_\alpha$ be the leaky-ReLU activation, $\alpha \in (0, 1)$,
and let $N \colon \R^d \to \R^d$ be a vanilla feedforward leaky-ReLU
network of depth $L$, with all hidden layers of width exactly $d$ and
non-singular weight matrices. Then there exists a Lipschitz
piecewise-affine vector field 
$v \colon \R^d \times [0, L] \to \R^d$, piecewise constant in $t$ on
each unit interval $[k, k+1)$ and computable in time 
$\mathrm{poly}(\mathrm{size}(N))$ from the parameters of $N$, such 
that $N$ is exactly the time-$L$ output of a Lie-Trotter splitting 
discretization of the non-autonomous ODE $\dot x = v(x, t)$ at step 
size $1$. In particular, $N(x_0)$ equals the $L$-th iterate of the 
splitting scheme started at $x_0$.
\end{lemma}
 
\begin{proof}
For each layer $\ell \in \{1, \dots, L\}$, the layer
$x \mapsto \sigma_\alpha(W_\ell x + b_\ell)$ is the Lie-Trotter
composition (\cite{trotter1959product}, see e.g.\ \cite{holden2010splitting}) of the time-$1$ flow of an affine ODE 
$\dot x = (W_\ell - I) x + b_\ell$ (whose flow at $t=1$ realizes 
$x \mapsto W_\ell x + b_\ell$ when $W_\ell$ is non-singular, after 
re-parameterization; see~\citealp{yifei2024vanilla} 
Section~3.5--3.6) and the time-$1$ flow of the activation ODE 
$\dot x = (\sigma_\alpha - I)(x)$ rescaled appropriately. Setting
$v(x, t) = v_\ell(x)$ for $t \in [\ell - 1, \ell)$, where $v_\ell$
encodes both ingredients of layer $\ell$, gives the required 
non-autonomous vector field. Composing $L$ unit splitting steps 
reproduces exactly $N$.
\end{proof}

\paragraph{From ReLU to leaky-ReLU at width $+1$.}
Lemma~\ref{thm:duan-correspondence} requires leaky-ReLU rather than
ReLU. The two activations are related by the elementary identity
\(
   \ReLU(x) \;=\; \dfrac{\sLR(x) - \alpha\,x}{1 - \alpha},
\)
which expresses ReLU as a linear combination of one leaky-ReLU
neuron and one identity register. By keeping $x$ alongside $\sLR(x)$
in a single extra register, every $\ReLU$ neuron at width $w$ can be
emulated by a leaky-ReLU subnetwork at width $w + 1$, with the same
depth modulo a constant. We record this as a folklore lemma.
 
\begin{lemma}[ReLU $\to$ leaky-ReLU emulation at width $+1$]
\label{lem:relu-to-leakyrelu}
For every fixed $\alpha \in (0, 1)$ and every $\ReLU$ network
$N \in \Netnotation{\ReLU}{n}{n}{w}$, there exists a leaky-ReLU network
$\widetilde N \in \Netnotation{\sLR}{n}{n}{w+1}$ with the same depth up
to a constant factor, such that $\widetilde N(x) = N(x)$ for all
$x \in \R^n$.
\end{lemma}

\begin{proof}
For each $\ReLU$ neuron $y = \ReLU(z)$ in $N$, the identity
$\ReLU(z) = (\sLR(z) - \alpha z)/(1-\alpha)$ allows us to compute $y$
as a linear combination of $\sLR(z)$ and $z$. Implementing this in a
vanilla leaky-ReLU FNN requires keeping $z$ alongside $\sLR(z)$
between layers; one extra register suffices for this passthrough,
where the identity wire is itself realized by a single leaky-ReLU
neuron via Proposition~\ref{kidger41}. The depth increases by a constant per
layer.
\end{proof}

\subsection{Neural ODE universality at dimension 5}
\label{sec:neural-ode-thm4}
 
Combining Theorem~\ref{thm:main} (width-$4$ ReLU universality) with
Lemma~\ref{lem:relu-to-leakyrelu} (ReLU $\to$ leaky-ReLU at width $5$)
and Lemma~\ref{thm:duan-correspondence} (leaky-ReLU width $d$
$\leftrightarrow$ non-autonomous ODE in dimension $d$) yields the 
following.
 
\begin{theorem}[Discretized Neural ODE consequence at dimension $5$]
\label{thm:neural-ode}
For every length bound $n \ge 1$, there exists a polynomial-time
constructible non-autonomous Lipschitz piecewise-affine vector field
$f_n \colon \R^5 \times [0, +\infty) \to \R^5$, piecewise constant in
$t$ on each unit interval and realizable layer-by-layer as a 
leaky-ReLU vanilla neural network of width $5$, and a fixed convex 
acceptance region $\AccODE \subseteq [0,1]^5$ specified by rational 
linear inequalities, such that the following holds. For every 
deterministic Turing machine $M$ over $\{0,1\}$ and every input 
$w \in \{0,1\}^*$ with $|w| \le n$, there is a 
polynomial-time-computable initial condition 
$\xi_0(M, w) \in [0,1]^5$ and an integer step count 
$L_M = \mathrm{poly}(|M|)$ such that the Lie-Trotter splitting 
discretization sequence $\{\xi_k\}_{k \ge 0}$ of 
$\dot \xi = f_n(\xi, t)$ at step size $1$, started at 
$\xi_0(M, w)$, satisfies:
\(
   M \text{ accepts } w
   \iff
   \exists\, k \ge 0,\quad
   \xi_{k \cdot L_M} \;\in\; \AccODE.
\)
Taking $M$ to be a fixed universal Turing machine $U$, $f_n$ becomes
a single non-autonomous vector field parameterized by length only, 
so the discretized flow simulates any $M$ on any input of length 
$\le n$.
\end{theorem}
 
\paragraph{Caveat on continuous flow.}
The discretization sequence $\{\xi_k\}$ is what reproduces the network
iteration; it is not in general equal to the continuous flow
$\xi \colon [0, +\infty) \to \R^5$ of $\dot \xi = f_n(\xi, t)$ sampled
at integer times. Splitting methods are only first-order accurate in
the step size, and for step size $1$ the gap is real. The 
continuous flow exists and is unique by Lipschitz of $f_n$, but its
values at integer times need not encode TM configurations. 
Furthermore, the vector field is non-autonomous (piecewise constant 
in $t$, with one piece per unit step); whether the construction can 
be carried out with an autonomous vector field is left open. See
\Cref{sec:neural-ode-open} for related open questions.
 
\begin{proof}
By Theorem~\ref{thm:main}, the width-$4$ $\ReLU$ network $N$ simulates
a universal Turing machine on convex inputs $[0,1]^4$. By
Lemma~\ref{lem:relu-to-leakyrelu}, an associated width-$5$
leaky-ReLU network $\widetilde N \colon [0,1]^5 \to [0,1]^5$ realizes
the same iteration on the embedded set
$\{(x, 0) : x \in [0,1]^4\} \subseteq [0,1]^5$. By
Lemma~\ref{thm:duan-correspondence} applied to $\widetilde N$, there
is a non-autonomous Lipschitz piecewise-affine vector field
$f_n \colon \R^5 \times [0, +\infty) \to \R^5$, piecewise constant in
$t$ on unit intervals, whose Lie-Trotter splitting discretization at
step $1$ over $L_M = \mathrm{poly}(|M|)$ time intervals reproduces
$\widetilde N$. Setting $\AccODE = \Acc \times \{0\}$ (with $\Acc$ 
from the verification recipe of \Cref{sec:zoo-proof}) closes the 
equivalence.
\end{proof}

\paragraph{Reading.}
Width $4$ in the discrete network and dimension $5$ in the Neural ODE
are the same quantity up to the leaky-ReLU emulation overhead. The
``$+1$'' is the cost of moving from $\ReLU$ to leaky-ReLU, the
splittable activation required by the Duan correspondence. The
question of whether \Cref{thm:main} can be carried out directly with
leaky-ReLU at width $4$ (which would yield a dimension-$4$ Neural
ODE), or whether the non-autonomous vector field can be made
autonomous, is left open.

\subsection{Verification hardness transfers}
\label{sec:neural-ode-hardness}
 
The verification spectrum of \Cref{thm:verif-spectrum} transfers to
Neural ODEs by Theorem~\ref{thm:neural-ode}.
 
\begin{corollary}[Verification spectrum for Neural ODEs at dimension $5$]
\label{cor:neural-ode-spectrum}
For Lipschitz piecewise-affine Neural ODEs $\dot \xi = f(\xi,t)$ in
dimension $5$, with convex initial and acceptance regions
$\InitODE, \AccODE \subseteq [0,1]^5$, the discretized reachability
problem ``does the splitting-discretized trajectory enter $\AccODE$
within $T$ discretization steps?'' is $\NP$-complete with $T$ unary,
$\NEXP$-complete with $T$ binary, and $\RE$-complete (undecidable)
without a time bound.
\end{corollary}
 
An analogous transfer of \Cref{thm:rho} (smooth $\rho$ at width $6$)
to Neural ODEs requires extending the
correspondence~\citep{yifei2024vanilla} from leaky-ReLU to other
activations. The Duan et al.\ argument is stated for leaky-ReLU and
relies specifically on its invertibility; we leave the extension to
$\tanh$, GELU, sigmoid, Swish, and other smooth activations as an open
direction.
 
\paragraph{Practical reading.}
A practitioner verifying a low-dimensional Neural ODE faces the full
hardness spectrum already at dimension $5$. The empirical observation
that complete verifiers degrade rapidly with integration time
\citep{katz2017reluplex,wang2021beta} is, on this account, an intrinsic
complexity-theoretic phenomenon and not a solver artifact: integration
time plays the role of Turing-machine time, and the hardness ladder is
unconditional on solver design.

\subsection{Open: pure continuous-time exact simulation}
\label{sec:neural-ode-open}
 
Theorem~\ref{thm:neural-ode} establishes universality of Neural ODEs
\emph{under splitting-method discretization}. The discretized flow at
integer times reproduces TM steps exactly. The continuous flow at
non-integer times is well-defined (by Lipschitz of $f$) but need not
encode any meaningful TM configuration.
 
A natural strengthening would be: a Lipschitz piecewise-affine
$f \colon \R^d \to \R^d$ whose \emph{continuous flow} satisfies
$\xi(n) = N^{[n]}(\xi_0)$ for every integer $n$, for $N$ a non-trivial
non-linear PA map. Whether this is achievable is open, and we record
two related questions.
 
\paragraph{Three known regimes.}
For the broader question of continuous-time TM simulation, three
regimes are established in the literature:
\begin{itemize}\setlength{\itemsep}{0pt}
\item Polynomial (rather than piecewise-affine) vector
fields~\citep{JOC2007} (analytic flow, but no longer in the Neural-ODE
class as commonly used).
\item Piecewise-constant discontinuous fields, e.g. PCD systems
\citep{AMP95,KCG94} (closed-form jumps at cell
boundaries, but not Lipschitz).
\item Splitting-method discretization (\Cref{thm:neural-ode}): exact
simulation at discretization times, with continuous interpolation
between them.
\end{itemize}
The Lipschitz piecewise-affine class, the class most aligned with
practical Neural ODEs, is not covered by any of these.
 
\paragraph{Two open questions.}
We highlight two natural open questions for future work, in increasing
order of strength:
\begin{itemize}\setlength{\itemsep}{0pt}
\item (Robust continuous-time) Does there exist an autonomous
Lipschitz piecewise-affine vector field $f \colon \R^d \to \R^d$
(for some constant $d$) such that the continuous flow simulates a
universal Turing machine in the following robust sense: for every
machine $M$ and input $w$, the trajectory enters an
$\varepsilon$-neighborhood of a designated accepting region in
finite time iff $M$ accepts $w$?
\item (Exact continuous-time) Does there exist such an $f$ with
$\xi(n) = N^{[n]}(\xi_0)$ exactly at integer times, for some
non-trivial PA map $N$ derived from $M$?
\end{itemize}
We leave both questions for future work.

\end{document}

%% file: FIG/fig-TM.tex
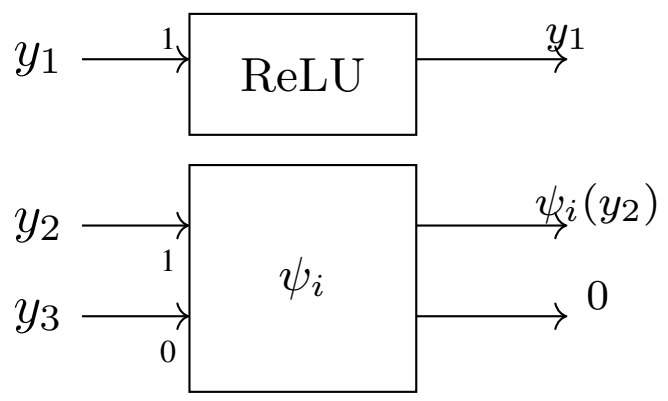
\begin{figure}
\begin{center}
\begin{tikzpicture}  

\node (y_1) at (0,0.5) {$y_{1}$};
\node (y_2) at (0,-0.6) {$y_{2}$};
\node (y_3) at (0,-1.2) {$y_{3}$};

\myrect{1}{0.8}{1.5}{-0.8}{$\ReLU$}
\myrect{1}{-0.2}{1.5}{-1.5}{$\psi_i$}

\draw[->] (y_1) to node[pos=0.8, yshift=4pt] {\tiny 1} (1,0.5);
\draw[->] (y_2) to node[pos=0.8, yshift=-1pt,below] {\tiny 1} (1,-0.6);
\draw[->] (y_3) to node[pos=0.8, yshift=-1pt,below] {\tiny 0} (1,-1.2);

\draw[->] (2.5,0.5)    -- node[pos=1, yshift=4pt] {\footnotesize $y_{1}$} (3.5,0.5);
\draw[->] (2.5,-0.6)  -- node[pos=1.2, yshift=4pt] { \footnotesize $\psi_i(y_{2})$}(3.5,-0.6);
\draw[->] (2.5,-1.2)  -- node[pos=1.2, yshift=4pt] { \footnotesize $0$}(3.5,-1.2);
\end{tikzpicture}  
\end{center}  
\caption{Width-$3$ block computing $(y_1, y_2, y_3)\mapsto(y_1,\psi_i(y_2),0)$.
Coordinates~$2$--$3$ host the width-$2$ subnetwork of Lemma~\ref{lem:cpam}
(staircase + shifted partial sum); coordinate~$1$ carries $y_1$ through each
layer via $\ReLU(y_1)=y_1$ on $[0,1]$. A final constant-depth junction outputs
$\psi_i(y_2)$ on coordinate~$2$ and resets coordinate~$3$ to~$0$.}

\label{fig:psi-block}
\end{figure}

%% file: FIG/fig-err.tex
\begin{figure}
\centering
\resizebox{\linewidth}{!}{%
\begin{tabular}{@{}c c@{}}
\adjustbox{valign=c}{%
\begin{tikzpicture}     
  
\node (x_in) at (0,0.5) {$x$};
\draw[->] (x_in) to node[pos=0.8, yshift=4pt] {\tiny 1} (1,0.5);
\draw[->] (x_in) to node[pos=0.8, yshift=5pt] {\tiny $-1$} (1,-0.5);
\myrect{1}{0.8}{1.5}{-0.8}{$\ReLU$}
\myrect{1}{-0.2}{1.5}{-0.8}{$\ReLU_{1}$}

\draw[->] (2.5,0.5) to node[pos=0.8, yshift=4pt] {\tiny 1} (3.5,0.5);
\draw[->] (2.5,0.5) to node[pos=0.8, yshift=5pt] {\tiny 1} (3.5,-0.5);
\draw[->] (2.5,-0.6) to node[pos=0.8, yshift=-1pt,below] {\tiny 1} (3.5,-0.6);

\myrect{3.5}{0.8}{1.5}{-0.8}{$\ReLU$}
\myrect{3.5}{-0.2}{1.5}{-0.8}{$\ReLU_{2}$}

 \draw[->] (5,0.5) to node[pos=0.8, yshift=4pt] {\tiny 1} (6,0.5);
\draw[->] (5,0.5) to node[pos=0.8, yshift=5pt] {\tiny -2} (6,-0.5);
\draw[->] (5,-0.6) to node[pos=0.8, yshift=-1pt,below] {\tiny -1} (6,-0.6);

\myrect{6}{0.8}{1.5}{-0.8}{$\ReLU$}
\myrect{6}{-0.2}{1.5}{-0.8}{$\ReLU_{5}$}

\draw[->] (7.5,0.5) to node[pos=0.8, yshift=4pt] {\tiny 1} (8.5,0.5);
\draw[->] (7.5,0.5) to node[pos=0.8, yshift=5pt] {\tiny -1} (8.5,-0.5);
\draw[->] (7.5,-0.6) to node[pos=0.8, yshift=-1pt,below] {\tiny -1} (8.5,-0.6);

 \myrect{8.5}{0.8}{1.5}{-0.8}{$\ReLU$}
\myrect{8.5}{-0.2}{1.5}{-0.8}{$\ReLU_{3}$}

\draw[->] (10,0.5) to node[pos=0.8, yshift=4pt] {\tiny $x$} (10.5,0.5);
\draw[->] (10,-0.6) to node[pos=1.2, yshift=4pt] {\tiny $\Err(x)$} (10.5,-0.6);

 \end{tikzpicture}  
 }
&
\adjustbox{valign=c}{%
\begin{tikzpicture}[scale=0.7]
  \draw[->] (-0.02,0) -- (4.2,0) node[right] {$x$};
  \draw[->] (0,-0.02) -- (0,2.5) node[above] {$\Err(x)$};

  \foreach \x/\label in {0/0,1/$1$,2/$2$,3/$3$,4/$4$}
    \draw (\x,0.05) -- (\x,-0.05) node[below=2pt] {\label};

  \foreach \y/\label in {1/0.5,2/1}
    \draw (0.02,\y) -- (-0.02,\y) node[left=2pt] {\label};

  \draw[dashed,gray] (0,1) -- (2.5,1);
 
  \draw[thick]   (0,2) -- (1,0) ;
  \draw[thick]   (1,0) -- (2,0) ;
  \draw[thick]   (2,0) -- (2.5,1) ;
  \draw[thick]   (2.5,1)  -- (3,0);
    \draw[thick]   (3,0) -- (4,0) ;

  \fill (0.25,1) circle (0.3pt);
  \fill (0.75,1) circle (0.3pt);
  \fill (0.75,3) circle (0.3pt);
  \fill (1,3) circle (0.3pt);
\end{tikzpicture}
}
\end{tabular}}
\caption{Network $\NNErr$.  Boxes represent $\ReLU$ neurons; subscripts indicate biases, edge
labels indicate weights, and missing labels correspond to bias \(0\) and weight~\(1\).}
\label{fig:err}
\end{figure}

%% file: FIG/fig-lead.tex
\begin{figure}
\begin{tabular}{@{}c c@{}}
\adjustbox{valign=c}{%
\begin{tikzpicture}  

\node (x_in) at (0,0.5) {$x$};

\draw[->] (x_in) to node[pos=0.8, yshift=4pt] {\tiny 1} (1,0.5);
\draw[->] (x_in) to node[pos=0.8, yshift=4pt] {\tiny $8$} (1,-0.5);
\myrect{1}{0.8}{1.5}{-0.8}{$\ReLU$}
\myrect{1}{-0.2}{1.5}{-0.8}{$\ReLU_{-6}$}

\draw[->] (2.5,0.5) to node[pos=0.8, yshift=4pt] {\tiny 1} (3.5,0.5);
\draw[->] (2.5,0.5) to node[pos=0.8, yshift=4pt] {\tiny 8} (3.5,-0.5);
\draw[->] (2.5,-0.6) to node[pos=0.8, yshift=-1pt,below] {\tiny -1} (3.5,-0.6);

\myrect{3.5}{0.8}{1.5}{-0.8}{$\ReLU$}
\myrect{3.5}{-0.2}{1.5}{-0.8}{$\ReLU_{-4}$}

 \draw[->] (5,0.5) to node[pos=0.8, yshift=4pt] {\tiny 1} (6,0.5);
\draw[->] (5,0.5) to node[pos=0.8, yshift=4pt] {\tiny 1} (6,-0.5);
\draw[->] (5,-0.6) to node[pos=0.8, yshift=-1pt,below] {\tiny 1} (6,-0.6);

 \myrect{6}{0.8}{1.5}{-0.8}{$\ReLU$}
\myrect{6}{-0.2}{1.5}{-0.8}{$\ReLU_{1}$}

  \draw[->] (7.5,0.5) to node[pos=0.8, yshift=4pt] {\tiny $x$} (8,0.5);
  \draw[->] (7.5,-0.6) to node[pos=1.2, yshift=4pt] {\tiny $\lead(x)$} (8,-0.6);

 \end{tikzpicture} %
 }
&
\adjustbox{valign=c}{%
\begin{tikzpicture}[scale=0.7]
  \draw[->] (-0.02,0) -- (4.08,0) node[right] {$x$};
  \draw[->] (0,-0.02) -- (0,3.25) node[above] {$\lead(x)$};

  \foreach \x/\label in {0/0,1/$\frac14$,2/$\frac12$,3/$\frac34$,4/1}
    \draw (\x,0.05) -- (\x,-0.05) node[below=2pt] {\label};

  \foreach \y/\label in {1/1,3/3}
    \draw (0.02,\y) -- (-0.02,\y) node[left=2pt] {\label};

  \draw[dashed,gray] (0,1) -- (2,1);
  \draw[dashed,gray] (0,3) -- (4,3);

  \draw[thick]   (0,1) -- (2,1) ;
   \draw[thick]      (2,1)  --  (3,3) -- (4,3);
   \draw[thick]     (3,3) -- (4,3);

  \fill (0.25,1) circle (0.3pt);
  \fill (0.75,1) circle (0.3pt);
  \fill (0.75,3) circle (0.3pt);
  \fill (1,3) circle (0.3pt);
\end{tikzpicture}
}
\end{tabular}
\caption{Network $\NNlead$.  Boxes represent $\ReLU$ neurons; subscripts indicate biases, edge
labels indicate weights, and missing labels correspond to bias \(0\) and weight~\(1\).}
\label{fig:lead}
\end{figure}

%% file: FIG/fig-Trunck-4.tex
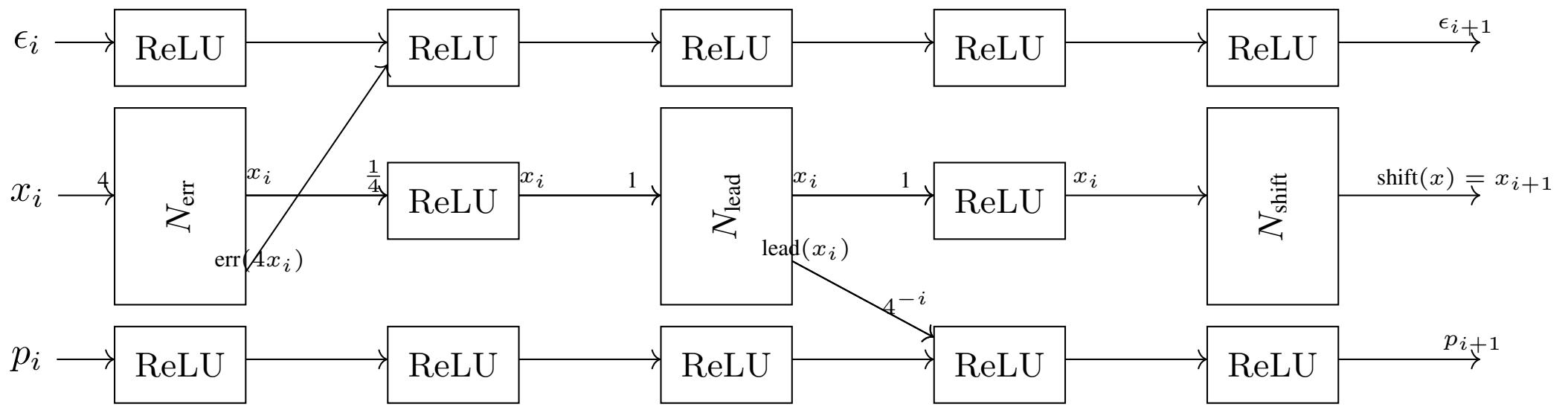
\begin{figure}
\centering
\resizebox{\linewidth}{!}{%
\begin{tikzpicture}
 \node (epsilon_in) at (-0.3,1.9) {$\epsilon_{i}$};   
\node (x_in) at (-0.3,0.5) {$x_{i}$};
\node (p_in) at (-0.3,-1) {$p_{i}$};

\draw[->] (x_in) to node[pos=0.8, yshift=4pt] {\tiny 4} (0.5,0.5);
\draw[->](epsilon_in) --  (0.5,1.9);   
\draw[->](p_in) --  (0.5,-1);

\myrectrotate{0.5}{1.3}{1.2}{-1.8}{$\NNErr$}
\myrect{0.5}{-0.7}{1.2}{-0.7}{$\ReLU$}
\myrect{0.5}{2.2}{1.2}{-0.7}{$\ReLU$}

 \draw[->] (1.7,0.5) to node[pos=0.1, yshift=5pt] {\tiny $x_{i}$} (3,0.5);

\draw[->] (1.7,0.5) to node[pos=0.9, yshift=5pt] {\tiny $\frac14$} (3,0.5);
\draw[->] (1.7,-0.2) to node[pos=0.1, yshift=4pt,below] {\tiny $\Err(4x_{i})$}  (3,1.7);

\draw[->](1.7,1.9) --  (3,1.9);   
\draw[->](1.7,-1) --  (3,-1);

\myrect{3}{-0.7}{1.2}{-0.7}{$\ReLU$}
\myrect{3}{2.2}{1.2}{-0.7}{$\ReLU$}
\myrect{3}{0.8}{1.2}{-0.7}{$\ReLU$}

 \draw[->] (4.2,0.5) to node[pos=0.8, yshift=4pt] {\tiny 1} (5.5,0.5);
 \draw[->] (4.2,0.5) to node[pos=0.1, yshift=4pt] {\tiny $x_{i}$ } (5.5,0.5);
\draw[->](4.2,1.9) --  (5.5,1.9);   
\draw[->](4.2,-1) --  (5.5,-1);

\myrectrotate{5.5}{1.3}{1.2}{-1.8}{$\NNlead$}
\myrect{5.5}{-0.7}{1.2}{-0.7}{$\ReLU$}
\myrect{5.5}{2.2}{1.2}{-0.7}{$\ReLU$}

\draw[->] (6.7,0.5) to node[pos=0.8, yshift=4pt] {\tiny 1} (8,0.5);
 \draw[->] (6.7,0.5) to node[pos=0.1, yshift=4pt] {\tiny $x_{i}$} (8,0.5);

\draw[->] (6.7,-0.1) to node[pos=0.8, yshift=5pt] {\tiny $4^{-i}$} (8,-0.8);
\draw[->] (6.7,-0.1) to node[pos=0.1, yshift=5pt] {\tiny $\lead(x_{i})$} (8,-0.8);
\draw[->](6.7,1.9) --  (8,1.9);   
\draw[->](6.7,-1) --  (8,-1);

\myrect{8}{2.2}{1.2}{-0.7}{$\ReLU$}
\myrect{8}{0.8}{1.2}{-0.7}{$\ReLU$}
\myrect{8}{-0.7}{1.2}{-0.7}{$\ReLU$}

\draw[->] (9.2,0.5) to node[pos=0.15, yshift=4pt] {\tiny $x_{i}$} (10.5,0.5);
\draw[->](9.2,1.9) --  (10.5,1.9);   
\draw[->](9.2,-1) --  (10.5,-1);      
 \myrect{10.5}{2.2}{1.2}{-0.7}{$\ReLU$}
 \myrect{10.5}{-0.7}{1.2}{-0.7}{$\ReLU$}
 \myrectrotate{10.5}{1.3}{1.2}{-1.8}{$\NNshift$}
 
\draw[->] (11.7,-1) to node[pos=0.95, yshift=4pt] {\tiny $p_{i+1}$} (13,-1);
\draw[->] (11.7,1.9) to node[pos=0.9, yshift=4pt] {\tiny $\epsilon_{i+1}$} (13,1.9);
  \draw[->] (11.7,0.5) to node[pos=0.9, yshift=4pt] {\tiny $\shift(x)=x_{i+1}$} (13,0.5);

\end{tikzpicture}}
\caption{Sub-network of $\NNT{n}$ implementing the $i$-th loop iteration, which updates $(x_i,p_i,\epsilon_i)$ into $(x_{i+1},p_{i+1},\epsilon_{i+1})$ using ReLU neurons and the three components $\NNErr$, $\NNlead$, and $\NNshift$. Outputs are shown at the sources of the arrows, and weights at their targets. The sub-network has width $4$. The components $\NNlead$ and $\NNshift$ have constant depth. For readability, only one ReLU layer is drawn in parallel with these components; formally, one should include as many ReLU layers as the depths of the corresponding sub-networks.}

\label{fig:SKIP-Trunck-4}
\end{figure}

%% file: FIG/fig-register.tex
\begin{figure}
\centering
\begin{tikzpicture}[>=stealth]

\node (x1) at (0, 1.5) {$x_{1}$};
\node (x2) at (0, 0.5) {$x_{2}$};
\node (x3) at (0,-0.5) {$x_{3}$};
\node (x4) at (0,-1.5) {$x_{4}$};

\draw[thick, rounded corners] (1,2) rectangle (3.5,-2);
\node[align=center] at (2.25, 0)
  {$N_{\mathrm{check}}$ \\[2pt] \footnotesize (width $4$)};

\draw[->] (x1) -- (1, 1.5);
\draw[->] (x2) -- (1, 0.5);
\draw[->] (x3) -- (1,-0.5);
\draw[->] (x4) -- (1,-1.5);

\draw[->] (3.5, 1.5) -- node[above, font=\footnotesize] {$x_{1}$}     (6, 1.5);
\draw[->] (3.5, 0.5) -- node[above, font=\footnotesize] {$x_{2}$}     (6, 0.5);
\draw[->] (3.5,-0.5) -- node[above, font=\footnotesize] {$x_{3}$}     (6,-0.5);
\draw[->] (3.5,-1.5) -- node[above, font=\footnotesize]
                         {$\max\{x_{4},\Cert(x_{1},x_{2})\}$}          (9,-1.5);

\draw[thick, rounded corners] (6,2) rectangle (9,-1);
\node[align=center] at (7.5, 0.5)
  {$g_{M}$ \\[2pt] \footnotesize (width $3$)};
\node[font=\footnotesize] at (7.5,-0.7) {scratch on coord.~$3$};

\draw[->] (9, 1.5) -- node[above, font=\footnotesize] {$x_{1}'$} (10.7, 1.5);
\draw[->] (9, 0.5) -- node[above, font=\footnotesize] {$x_{2}'$} (10.7, 0.5);
\draw[->] (9,-0.5) -- node[above, font=\footnotesize] {$0$}      (10.7,-0.5);
\draw[->] (9,-1.5) --                                            (10.7,-1.5);

\node[font=\footnotesize] at (7.5,-1.85) {identity wire};

\end{tikzpicture}
\caption{The width-$4$ network $N_M = N_{\mathrm{sim}}\circ N_{\mathrm{check}}$.
The block $N_{\mathrm{check}}$ updates the certificate coordinate~$x_4$
to $\max\{x_4,\Cert(x_1,x_2)\}$ while leaving $(x_1,x_2,x_3)$ untouched;
the block $N_{\mathrm{sim}}$ then applies the width-$3$ simulator
$g_M$ to $(x_1,x_2)$ with $x_3$ as scratch, and propagates the updated
$x_4$ through an identity wire.}
\label{fig:register}
\end{figure}
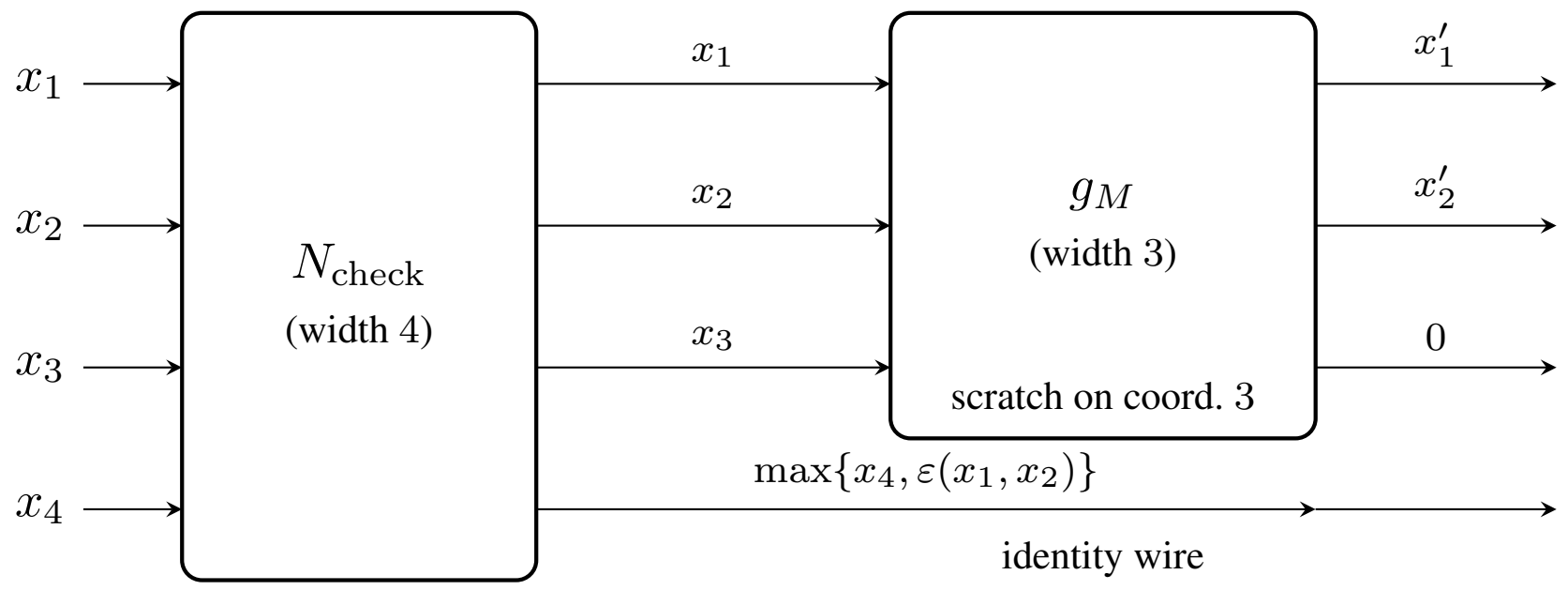